%% file: main.tex
\documentclass[10pt,twocolumn,letterpaper]{article}

\usepackage{cvpr}              

\usepackage{graphicx}
\usepackage{amsmath}
\usepackage{amssymb}
\usepackage{booktabs}
\usepackage{float}
\usepackage{booktabs}
\usepackage{color}
\usepackage[dvipsnames,table,xcdraw]{xcolor}
\def\method{ViiNeuS}
\definecolor{sdf}{RGB}{230, 255, 230}
\definecolor{mvs}{RGB}{230, 240, 255}
\usepackage{overpic}
\usepackage{multirow}
\usepackage{stfloats}
\usepackage{soul}
\usepackage{wrapfig}
\usepackage{graphicx}
\usepackage{pict2e}

\definecolor{cvprblue}{rgb}{0.21,0.49,0.74}
\usepackage[pagebackref,breaklinks,colorlinks,allcolors=cvprblue]{hyperref}

\def\paperID{13257} 
\def\confName{CVPR}
\def\confYear{2025}

\title{\method: Volumetric Initialization for Implicit Neural Surface reconstruction of urban scenes with limited image overlap}

\author{
    Hala Djeghim$^
    {1,2}$ \qquad
    Nathan Piasco$^{1}$ \qquad
    Moussab Bennehar$^{1}$ \qquad
    Luis Roldão$^{1}$ \\
    Dzmitry Tsishkou$^{1}$ \qquad
    Désiré Sidibé$^{2}$  \\
    $^{1}$Noah's Ark, Huawei Paris Research Center, France \\
    $^{2}$IBISC, Evry Paris-Saclay University, France\\
    }

\begin{document}
\input{figures_tex/teaser/teaser.tex}
\maketitle
\input{sec/0_abstract}    
\input{sec/1_intro}
\input{sec/2_related_work}

\input{sec/3_method}

\input{sec/4_experiments}

\input{sec/5_discussion}

\input{sec/6_conclusion}
\input{sec/X_suppl}

\newpage
\newpage

{
    \small
    \bibliographystyle{ieeenat_fullname}
    \bibliography{main}
}

\end{document}

%% file: figures_tex/teaser/teaser.tex
\twocolumn[{%
\renewcommand\twocolumn[1][]{#1}%
\maketitle

\begin{center}
	\centering
    \begin{overpic}[width=.89\linewidth]{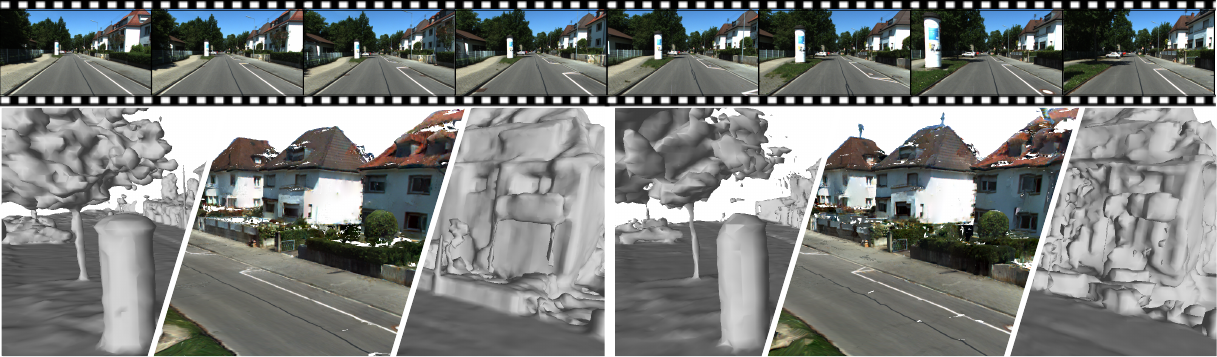}
    \put(21.5, -3){\method}
    \put(69, -3){StreetSurf~\cite{streetsurf}}
    \end{overpic}
  \vspace{2mm}
	\captionof{figure}{We introduce \textbf{\method}, a novel SDF initialization method tailored to accurately reconstruct large-scale driving scenes from RGB images with limited overlaps. Extensive experiments on popular driving datasets show the superiority of \method's mesh (left) over previous state-of-the-art methods such as StreetSurf\cite{streetsurf} (right). The figure presents both the initial mesh and the textured one as derived from both techniques.}
	\label{fig:teaser}
\end{center}
}]

%% file: sec/0_abstract.tex
\begin{abstract}

Neural implicit surface representation methods have recently shown impressive 3D reconstruction results. However, existing solutions struggle to reconstruct driving scenes due to their large size, highly complex nature and their limited visual observation overlap.
Hence, to achieve accurate reconstructions, additional supervision data such as LiDAR, strong geometric priors, and long training times are required.
To tackle such limitations, we present~\textbf{~\method}, a new hybrid implicit surface learning method that efficiently initializes the signed distance field to reconstruct large driving scenes from 2D street view images.
\method's hybrid architecture models two separate implicit fields: one representing the volumetric density of the scene, and another one representing the signed distance to the surface.
To accurately reconstruct urban outdoor driving scenarios, we introduce a novel volume-rendering strategy that relies on self-supervised probabilistic density estimation to sample points near the surface and transition progressively from volumetric to surface representation. 
Our solution permits a proper and fast initialization of the signed distance field without relying on any geometric prior on the scene, compared to concurrent methods.
By conducting extensive experiments on four outdoor driving datasets, we show that \method~can learn an accurate and detailed 3D surface representation of various urban scene while being two times faster to train compared to previous state-of-the-art solutions.
\end{abstract}

%% file: sec/1_intro.tex
\section{Introduction}
\label{sec:intro}
\label{sec:intro} 
Achieving accurate 3D reconstruction from multi-view-consistent images has been an important topic in computer vision and computer graphics communities for decades. A 3D representation of the perceived environment is fundamental for many tasks, such as scene re-lighting~\cite{zhang2023neilf}, scene editing~\cite{Zhu_2023_CVPR} and 3D object insertion for artificial data augmentation~\cite{fegr}.

Recently, neural implicit surface representation methods~\cite{NeuS, VolSDF} have shown impressive 3D surface reconstruction performance from monocular images, achieving highly detailed and precise mesh representations.
Such methods are originally inspired by Neural Radiance Fields (NeRF)~\cite{2020nerf}, and re-framed for multi-view surface reconstruction.
NeRF introduced volume rendering for Novel View Synthesis (NVS) by sampling points along camera rays. These points, along with their viewing directions, are fed into a Multi-Layer Perceptron (MLP) to predict their respective colors and densities. The final image is rendered by blending these predicted values using alpha compositing.
Conversely, neural surface representation methods aim to approximate the surface using a learnable Signed Distance Function (SDF)~\cite{NeuS, VolSDF}.
Unlike NeRF, the outputs of the MLP are colors and SDF values instead of colors and volumetric densities, which are later blended using a modified alpha-compositing equation adapted to SDF.
An accurately reconstructed surface is represented as the zero-level set of the SDF. Assuming objects are centered in the scene, the SDF is initially set to a spherical shape to bound the region of interest. While current methods achieve impressive reconstruction results, they are mainly designed for bounded, high-overlap observations of object-centric or landmark scenes. Applying them to urban scenes with limited image overlap leads to sub-optimal reconstructions due to their significantly different nature. Challenges include partial scene observability (e.g., occlusions from parked cars) and the need to handle both close- and far-range objects. Moreover, the geometry of driving scenes is highly complex, with fine structures and large untextured areas like roads and buildings.
To overcome such limitations, implicit surface reconstruction of large outdoor driving scenes commonly relies on LiDAR supervision~\cite{urban-radiance-fields, fegr} or strong geometric priors~\cite{streetsurf}. These approaches require longer training time to produce qualitative outdoor scene reconstruction meshes~\cite{li2023neuralangelo}.

To address those challenges, we propose \method, a hybrid approach to accurately initialize a signed distance field of a large area and represent complex surfaces from urban outdoor driving scenarios with high accuracy and geometric precision while requiring half of the training time compared to other state-of-the-art solutions~\cite{streetsurf}.
We propose a hybrid architecture that models two separate implicit fields: one representing the volumetric density of the scene and another one representing the signed distance to the surface. 
\method's novel volume rendering approach relies on a self-supervised density estimator which permits to sample points near the surface along with a progressive sampling strategy that smoothly initializes and transitions from a volumetric to a surface representation, drastically reducing the model convergence time.
Inspired by previous state-of-the-art works~\cite{streetsurf,Yu2022MonoSDF}, we supervise surface normals by relying on normal cues predicted from a pre-trained monocular estimator \cite{eftekhar2021omnidata}.
We extensively validate our solution on KITTI-360~\cite{Kitti}, Pandaset~\cite{pandaset}, Waymo Open Dataset~\cite{Waymo}, and nuScenes~\cite{nuscenes}. By comparing the quality of the extracted meshes from the signed distance field, we show that our method can efficiently reconstruct urban outdoor scenarios showing SoTA accuracy. 

\noindent To summarize, the main contributions of our work are:
\begin{itemize}
       \item a novel learning method that progressively initialize a signed distance field from street views with limited images overlap,
        \item a hybrid architecture, suited to the complexity of outdoor urban scene, that models two distinct implicit fields to represent both the volumetric density and the signed distance field,
        \item a new volume rendering technique that relies on self-supervised probabilistic density estimation to efficiently sample points near the surface, significantly reducing overall training times to obtain a scene mesh,
        \item an extensive comparison of SoTA implicit SDF and explicit MVS reconstruction solutions on 4 major driving datasets, showing the adaptability and superiority of \method~for urban scenes surface reconstruction.
    \end{itemize}

%% file: sec/2_related_work.tex
\section{Related work}

\noindent\textbf{Multi-view surface reconstruction.}
Traditional surface reconstruction techniques represent a scene either by occupancy grids~\cite{de1999poxels, paschalidou2019raynet, tulsiani2017multiview, 7335464} or relying on point clouds~\cite{galliani2016gipuma, schonberger2016pixelwise, stereopsis2010accurate, Roldo20193DSR}. Occupancy grids-based methods are often limited by the voxel resolution and suffer from high memory consumption. 
On the other hand, Multi-View Stereo (MVS) techniques~\cite{openMVG} generate, from posed images, dense point clouds associated with surface normal that can be utilized in a subsequent surface reconstruction step~\cite{kazhdan2006poisson}.
These methods heavily depend on the quality of the generated point cloud. Nevertheless, they can accumulate errors due to their complex multi-step pipelines, and are mostly designed for object-centric scenes with high overlaps between the observations. We refer the reader to dedicated surveys for in-depth analysis of such techniques \cite{Zollhfer2018StateOT, Malleson20193DRF, Berger2017ASO}.\\

\noindent\textbf{Neural implicit surface representations.} 
NeRF-based methods achieve impressive NVS results but fail to accurately reconstruct scene surfaces given the lack of constraints or priors on the surface when modeling the scene as a density volume.
Neural surface reconstruction approaches, explicitly represent the surface as an SDF and propose various adaptations to train such methods through differentiable volume rendering~\cite{NeuS, VolSDF}.
NeuS~\cite{NeuS} proposes a novel SDF-based rendering equation that relies on SDF values to compute the alpha blending factor used in volume rendering instead of densities as in NeRF~\cite{2020nerf}.
However, NeuS's geometric initialization is designed for object-centric datasets and requires long training times. Recent works aim to reduce the training time~\cite{neus2} or reconstruct large-scale highly detailed landmark scenes~\cite{li2023neuralangelo}. Neuralangelo~\cite{li2023neuralangelo}, a SoTA method for outdoor landmarks reconstruction, requires many overlapping images assuming the structure to be reconstructed is centered within the scene. Indeed, similar to NeuS, they initialize the SDF with a spherical shape.
Such approaches are not tailored for large unbounded driving scenarios and still require hours of training. In contrast, our method is built upon NeuS's formulation but is designed to reliably represent urban outdoor driving scenes.\\
\noindent\textbf{Driving scenes surface reconstruction.}
Few surface reconstruction methods consider acquisition from a moving camera rig for driving sequences surface reconstruction.
Existing solution are mainly guided by LiDAR such as FEGR~\cite{fegr} and Urban Radianc Fields~\cite{urban-radiance-fields}, limiting their large-scale deployment.
StreetSurf~\cite{streetsurf}, the most related method to our proposal, divides the scene into separate close-range and far-range views and uses ego pose-based road surface initialization. This method assumes the road and sidewalk are at the same height across the entire scene, which limits its applicability to long, narrow driving scenarios, as seen in the Waymo Open Dataset~\cite{Waymo}. Instead, \method~does not rely on such topological assumptions, making it more adaptable to a variety of driving scenes, including those that are short and curvy. Additionally, \method~is more compact and faster to train, as it represents the entire scene with a single hybrid implicit field.\\

\noindent\textbf{Hybrid scene representation.}
Some state-of-the-art methods have explored hybrid rendering solutions combining volumetric and surface representations. Wang et al.~\cite{wang2023adaptive} introduced a hybrid method based on NeuS~\cite{NeuS} using a spatially varying kernel that adapts based on surface fuzziness. This kernel allows the extraction of a narrow mesh envelope around the surface, which enables ray casting only inside this envelope, accelerating both rendering time and quality.
Other methods~\cite{VMesh, huang2024sur2f, tang2023delicate} combine explicit mesh representations with SDF representations. The hybrid volume-mesh representation reduces storage and resource demands while outputting implicit geometry. Additionally, methods like NeuSG~\cite{chen2023neusg} and other 3D Gaussian Splatting-based~\cite{kerbl3Dgaussians} solutions~\cite{Guedon_2024_CVPR, Huang2DGS2024} aim to optimize Gaussians for surface reconstruction, with some approaches directly identifying level sets for surface extraction, such as in Gaussian Opacity Fields~\cite{Yu2024GOF}. 
The aforementioned solutions mainly focus on improving the rendering speed and are better fitted for scenes with high image overlap. In our proposal, we jointly train a hybrid model to benefit from the fast training of a density-based approach along with an accurate geometric representation obtained thanks to the signed distance field learning.

%% file: sec/3_method.tex
\section{Method}

\input{figures_tex/scilla_overview/scilla_overview.tex}

\method~models the volumetric density and signed distance field in a hybrid architecture.
The details of this architecture are presented in Sec.~\ref{sec:architecture}, while the novel underlying volume rendering strategy is explained in Sec.~\ref{sec:SCILLA-vol-rendering}.
Our hybrid strategy relies on a self-supervised density distribution estimation to sample points near the surface and progressively transition from a volumetric to an implicit surface representation from limited overlapping street views. To facilitate the convergence of our method, we further introduce regularization techniques for the signed distance field.
The overall architecture of our proposed method is illustrated in Fig.~\ref{fig:method-overview}.

\subsection{Preliminaries}

\paragraph{NeRF's Implicit Volumetric Representation.} 
 \label{paragraph:nerf}
 NeRF \cite{2020nerf} takes as input the 3D positions $x_i \in \mathbb{R}^3$ and the viewing direction $d \in \mathbb{R}^2$ of sampled points along each camera ray $r$ and predicts their corresponding densities $\sigma_i \in \mathbb{R}$ and colors $c_i \in \mathbb{R}^3$. This representation of the scene is approximated using a Multi-Layer Perceptron (MLP). Finally, volume rendering is used to alpha-composite the colors of the samples along each ray and render the final pixel color $\hat{C}(r) \in \mathbb{R}^3$:

\begin{equation}
    \hat{C}(r) = \sum_{i=1}^{N} w_i c_i ,  \ \  w_i = T_i \alpha_i, 
    \label{eq:volume-rendering}
\end{equation}
where $T_i = \prod_{j=1}^{i-1} (1- \alpha_j)$ is the accumulated transmittance along the ray and $\alpha_i \in \mathbb{R}$ a blending factor. This factor (referred to in the literature as the alpha-value) is used for alpha-composing, and is given by:
\begin{equation}
    \alpha_i = 1 - \exp (-\sigma_i \delta_i ),
    \label{eq:alpha-volume}
\end{equation}
with $\delta_i \in \mathbb{R}$ being the distance between the samples along the ray. %
Note that in \method's hybrid formulation, we refer to this value as $\alpha^{v}_i$ to disambiguate from alpha values derived from the estimated SDF which is introduced in Eq.~\ref{eq:neus-alpha}.

\paragraph{SDF representation with NeuS.} \label{sec-neus}
For an accurate surface-based representation of a 3D scene, NeuS \cite{NeuS} presented a different formulation of the NeRF's volume rendering equation.
They represent the scene as a signed distance function, where the surface is the zero-level set of such function:
$
    S = \left \{ x \in \mathbb{R}^3 | f(x) = 0\right \}
$, 
where $f(x)$ is the signed distance value at a given position $x$.
NeuS adopts a volume rendering formulation that introduces a density distribution function $\phi_s (x)$, which is formulated as the derivative of the sigmoid function $\Phi_s(x) = (1 + e^{-sx})^{-1}$, that is,
$ \phi_s (x) = se^{-sx} / (1 + e^{-sx})^2$.
The standard deviation of $\phi_s$ is $1/s$ and, as the network converges to an accurate surface representation, this parameter moves toward zero.
Following NeuS's formulation, the volumetric alpha used in Eq.~\ref{eq:volume-rendering} is replaced by the following equation to approximate the color $\hat{C}(r)$: 

\begin{equation}
    \alpha_i = \text{max} \left(\frac{ \Phi_s(f(p_i)) - \  \Phi_i(f(p_{i+1}))}{ \Phi_s(f(p_i))}, 0\right),
    \label{eq:neus-alpha}
\end{equation}
\noindent where $f(p_i)$ and $f(p_{i+1})$ are signed distance values at section points centered on $x_i$.
In our formulation, we denote alpha values derived from the estimated SDF as $\alpha_i^{f}$.

\paragraph{SDF field initialization}
In~\cite{yariv2020multiview}, the authors show that geometric initialization of the signed distance field (SDF) to a bounded sphere is key for learning an SDF function. However, such initialization assumes full scene observability from the training views and a centered region of interest, which does not apply to driving scenes where the camera mounted on the ego-vehicle remains on the road and follows a linear trajectory. StreetSurf~\cite{streetsurf} initializes the SDF based on road-surface. Instead, we propose a progressive initialization of the SDF, tailored for the complexities of urban scenes, ensuring effective SDF learning across both close and far-range elements in the entire driving sequence.

\subsection{\method's hybrid architecture}

\label{sec:architecture}
Our hybrid method \method~learns the two different fields simultaneously, as illustrated in Fig~\ref{fig:scilla_archi}. \method~expresses the scene using two functions, $\mathcal{F}_\Theta^h$ and $\mathcal{F}_\Theta^c$, encoded, respectively, by two MLPs. 
The input 3D position $x$ and the viewing direction $d$ are embedded, respectively, through a positional and directional embedding into $\text{emb}_x(x)$ and $\text{emb}_d(d)$.\\
$\mathcal{F}_\Theta^h$ takes as input $\text{emb}_x(x)$ and outputs a density $\sigma$, a signed distance value $f(x)$, and a latent vector $h$, that is 
$
    (\sigma, f(x), h) = \mathcal{F}_\Theta^h \left ( \text{emb}_x(x) \right ). %
$

\noindent $\mathcal{F}_\Theta^c$ on the other hand, takes as input the latent vector $h$ that is outputted by $\mathcal{F}_\Theta^h$ in addition to the embedded viewing direction $\text{emb}_d(d)$ and the normal vector $\overrightarrow{n}$ obtained from the SDF gradient, to predict the color $c = \mathcal{F}_\Theta^c \left ( \text{emb}_d(d), h, \overrightarrow{n} \right )$.

\input{figures_tex/scilla_architecture/scilla_archi.tex}

\subsection{\method's volume rendering}\label{sec:SCILLA-vol-rendering}

\label{subsec:SCILLA-samples-selection}
    \paragraph{Probabilistic density estimation.}
    \label{sec:density-estimation}
    Probabilistic density estimation along the ray is employed for a more targeted sampling strategy that samples fewer points and accelerates the overall method's training. Guided by the estimated density, $N$ points are sampled along the ray, where higher density indicates the presence of the surface. Using \method's architecture, alpha values are computed for these samples, and volume rendering is performed to obtain the final color associated with the ray:    
    \begin{equation}
        \begin{split}
        \hat{C}(r) & = \sum_{i=1}^{N} T_i \alpha_i c_i, \\
        \text{with~} \alpha_i & =
            \begin{cases}
              \alpha_i^v & \text{if computed from $\sigma_i$ (Eq.~\ref{eq:alpha-volume})},\\
              \alpha_i^f & \text{if computed from $f(x_i)$ (Eq.~\ref{eq:neus-alpha})}.
            \end{cases}      
        \end{split}
    \label{eq:scilla-volume-rendering}
    \end{equation}
    This resulting color is then compared to the ground truth value to train our model.
    The strategy of associating a sample to either a volumetric or surface-based representation is discussed in the next section.
    
    The weights $w_i$ associated with each sample (Eq.~\ref{eq:volume-rendering}) serve as a supervision signal to our probabilistic density estimator. Hence, we can refine in a self-supervised manner the prediction of the expected density along the ray. We rely on Mip-NeRF 360's~\cite{mip-nerf-360} proposal networks as our volumetric density estimator and use their introduced proposal loss, $\mathcal{L}_{\text{prop}}$, to refine density estimation from the weights $w_i$ of our main model.
  
\paragraph{SDF progressive initialization: Volumetric to surface representation.}
 \label{sec:vol2sdf}
Starting from a pure volumetric density training ($\alpha_i = \alpha_i^v$), we introduce a progressive initialization of the signed distance field that transitions from a volumetric to an SDF representation. Our approach is based on the assumption that estimating the density $\sigma_i$ at a specific location requires less complex geometric understanding, in contrast to estimating the signed distance to the closest point to a complex urban outdoor surface (i.e. $f(x_i)$). The 3D geometry represented as a density field contains less information than the SDF field; the former contains occupancy information (which can be considered as binary) whereas the latter represents, additionally, continuous distances to the surface,  including close and far-range points as well as the empty spaces between driving scene objects, such as cars and trees. Therefore, in this specific scenario where observations have reduced overlap introducing ambiguity in the geometry estimation, learning a density field is simpler compared to the SDF field (see figure~\ref{fig:vol_vs_sdf} of our ablation section). Furthermore, by initially leveraging a pure volumetric density representation during training, we shorten the convergence time of our density estimator.

\noindent Our progressive initialization strategy can be outlined as follows:
\begin{itemize}
    \item Volumetric stage: All samples along the ray express volumetric densities, and their alpha values are computed with $\sigma_i$.
    \item Hybrid stage: we gradually assign samples along the ray to train our signed distance field. Within the same ray, a portion of the sample values is represented by volumetric alpha $\alpha_i^v$, while the remainder is computed with $\alpha_i^{f}$.
    \item Surface stage: we increase the number of samples for which alpha values are computed using $f(x)$ until we transition to a complete surface representation.
\end{itemize}

\paragraph{Hybrid stage regularization.}
\label{sec:sd-reg}
Composing alpha values obtained from our two different geometric representations is not straightforward as the learning dynamic of the density and the SDF are not the same. As mentioned earlier and due to the partial observability of the reconstructed environment, learning a volumetric representation of the scene based only on spatial density is easier and faster compared to learning an arbitrary signed distance field of a complex urban scene. To facilitate the convergence of our representation, we introduce two regularization methods: \textbf{probability-based sampling attribution} and \textbf{SDF-gradient normalization}.\\
\noindent\textit{Probability-based samples attribution.} To assign the alpha value based on density or SDF to a sample, we initially start with a random selection for each ray. We noticed that this association strategy is not suitable: the signed distance field could not be properly initialized during the hybrid stage. To solve this problem, we introduce a probability-based sample association. We assign $\alpha_{i}^f$ to samples with the highest predicted density to initialize the signed distance field. It facilitates the learning of the signed distance field by limiting the range of predicted values, as all the points are sampled close to a surface. By sampling points near the surface, we handle the wide range of distances of driving scenarios and complex geometries.\\
\noindent\textit{SDF-gradient normalization.} 
During the hybrid stage, where $\alpha_i^v$ and $\alpha_i^f$ are jointly composed, we observed that the surface representation often predicts large SDF gradient values to align $\alpha_i^f$ with $\alpha_i^v$ (for more details, see our supplementary materials). To address this, we normalize the SDF gradient in the $\alpha_i^f$ computation to prevent the gradient from compensating the alpha distribution difference.

\subsection{Optimization}

\noindent\textbf{Photometric loss.} We use the standard $L_1$ loss to minimize the pixel-wise color difference between the rendered image $\hat{C}$, and the ground truth image $C$, as well as DSSIM~\cite{wang2023planerf} loss on color patches.\\

\noindent\textbf{Monocular cues.} \label{sec:mono-loss}
As we target to reconstruct outdoor datasets, we add an additional sky loss to our model. Similar to StreetSurf~\cite{streetsurf}, we model the sky color with an auxiliary MLP conditioned on the ray direction and use a loss to constrain the opacity of sky pixels to zero. The segmentation mask is obtained with an off-the-shelf semantic segmentation network~\cite{cheng2021mask2former}.

Inspired by MonoSDF\cite{Yu2022MonoSDF} and StreetSurf \cite{streetsurf}, we use off-the-shelf monocular depth estimation models to supervise the rendered normals $\hat{N}(r)$.
In standard solutions \cite{Yu2022MonoSDF,streetsurf}, the monocular normal loss is minimized according to the predicted normal accumulated along the ray:
\begin{equation}
    \hat{N}(r) = \sum_{i=1}^{N} T_i \alpha_i \frac{\nabla f(x_i)}{\left \|\nabla f(x_i)\right \|_2} ,
\end{equation}
\begin{equation}
    \mathcal{L}_{\hat{N}} = \left \|  \hat{N}(r) - \bar{N}(r)\right \| + \left \| 1 - \hat{N}(r)^\top  \bar{N}(r) \right \|.
\end{equation}

In our proposal, however, for more accurate and targeted supervision, we supervise only the normal associated with the closet sample $x_N$ to the surface estimated by transmittance saturation:

\begin{multline}
        \mathcal{L}_{\hat{N}} = \left \|  \frac{\nabla f(x_N)}{\left \|\nabla f(x_N)\right \|_2} - \bar{N}(r)\right \| + \\
        \left \| 1 - \left ( \frac{\nabla f(x_N)}{\left \|\nabla f(x_N)\right \|_2} \right ) ^\top  \bar{N}(r) \right \|.
\end{multline}

\noindent\textbf{Signed distance field regularization.} \label{sec:sdf_reg_loss}
Similar to~\cite{NeuS}, to constrain our model to respect the eikonal equation~\cite{icml2020_2086} we use the eikonal regularization defined as:
\begin{equation}
    \mathcal{L}_{\text{eik}} = \frac{1}{N} \sum_{i=1}^{N} \left( \left \| \nabla f(x_i) \right \|_2 - 1 \right)^2.
\end{equation}

According to NeuS formulation of the alpha compositing equation (Eq.~\ref{eq:neus-alpha}), the network converges properly when the standard deviation $1/s$ of $\phi_s$ approaches zero. Due to the complexity of driving scenarios and the large scale of urban scenes, the parameter $s$ doesn't always reach a satisfactory level of convergence. Unlike StreetSurf which manually fixes the parameter $s$ to increment linearly, we introduce a regularization term on $s$ during training: $\mathcal{L}_{\text{s}} = 1 / (s + \epsilon)$.

\input{tables_tex/main_quantitative_tab.tex}

%% file: figures_tex/scilla_overview/scilla_overview.tex
\begin{figure*}[!ht]
    \centering
    \begin{overpic}[width=0.88\textwidth]{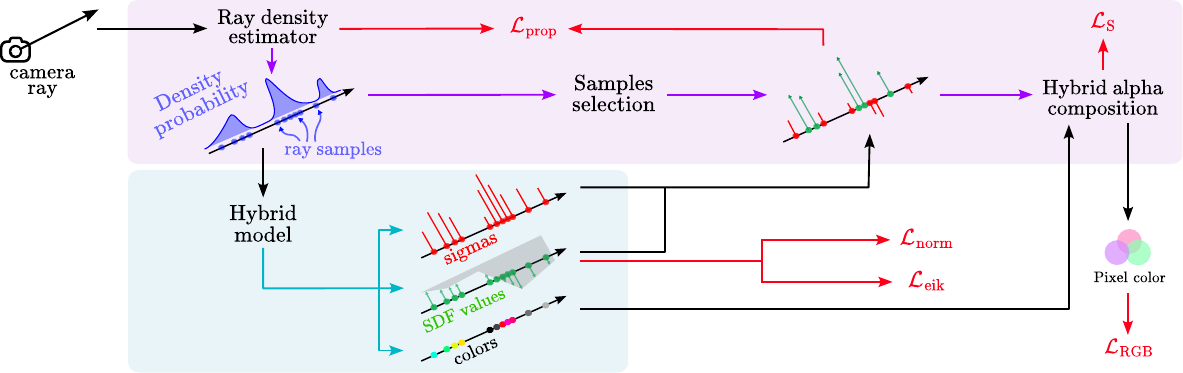}
        \put(13.5, 4.7){\footnotesize \method~architecture}
        \put(17.6, 2.5){\footnotesize (Sec. \ref{sec:architecture})}
        \put(34.4, 18.60){\footnotesize \method~Volume Rendering (Sec. \ref{sec:SCILLA-vol-rendering})}
    \end{overpic}
    \caption{\textbf{\method~ overview --} Our solution can be divided into two key components: an hybrid scene representation (Sec. \ref{sec:architecture}) and a ray-based volumetric rendering that progressively transitions from density to SDF sample alpha composition (Sec. \ref{sec:SCILLA-vol-rendering}).}
    \label{fig:method-overview}
    \hfill
    \vspace{-1mm}
\end{figure*}

%% file: figures_tex/scilla_architecture/scilla_archi.tex
\begin{figure}[!t]
\centering
\includegraphics[width=0.89\linewidth]{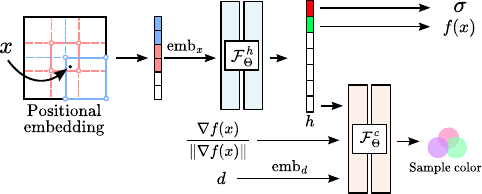} 
\caption{\textbf{\method~ architecture -- } 
Our method uses two MLP functions $\mathcal{F}_\Theta^h$ and $\mathcal{F}_\Theta^c$ to output SDF values and density ($\sigma$ and $f(x)$, respectively) along with color values given an input sample. We inspire from NeRF and design $\mathcal{F}_\Theta^c$ to output the color given the lattent vector $h$ outputted by $\mathcal{F}_\Theta^h$, the viewing direction ($d$) and the normal vector $\overrightarrow{n}$ obtained from the gradient of the SDF.}
\label{fig:scilla_archi}
\end{figure}

%% file: tables_tex/main_quantitative_tab.tex
\begin{table*}[tb]
    \centering
    \resizebox{1.7\columnwidth}{!}{%
    \begin{tabular}{lcccccccccccccccccc}
    
    & \multicolumn{8}{c}{KITTI-360~\cite{Kitti}}  
    & & \multicolumn{8}{c}{Pandaset~\cite{pandaset}}\\
    \cmidrule[\lightrulewidth]{2-9}
    \cmidrule[\lightrulewidth]{11-18}
    & \multicolumn{2}{c}{Seq. 30} & \multicolumn{2}{c}{Seq. 31} & \multicolumn{2}{c}{Seq. 35} & \multicolumn{2}{c}{Seq. 36} & & 
    \multicolumn{2}{c}{Seq. 23} & \multicolumn{2}{c}{Seq. 37} & \multicolumn{2}{c}{Seq. 42} & \multicolumn{2}{c}{Seq. 43}\\
    \cmidrule(lr){2-3} \cmidrule(lr){4-5} \cmidrule(lr){6-7}\cmidrule(lr){8-9}
    \cmidrule(lr){11-12} \cmidrule(lr){13-14} \cmidrule(lr){15-16}\cmidrule(lr){17-18}
    & P$\rightarrow$M & Prec.  & P$\rightarrow$M & Prec. & P$\rightarrow$M & Prec. & P$\rightarrow$M~ & Prec. &
    & P$\rightarrow$M & Prec. & P$\rightarrow$M & Prec. & P$\rightarrow$M & Prec. & P$\rightarrow$M & Prec.  \\
    \cmidrule[\lightrulewidth]{1-9}
    \cmidrule[\lightrulewidth]{11-18}

    StreetSurf~\cite{streetsurf} & 0.14 & 0.50 &\cellcolor{sdf}  \textbf{ 0.09} & \cellcolor{sdf} \textbf{0.71} &   \cellcolor{sdf} \textbf{0.10} &   0.67 &  \cellcolor{sdf} \textbf{0.11}  & 0.66 & & 2.33 & 0.10 & 0.47 & 0.40 & 0.36& 
  0.30& \cellcolor{sdf}  \textbf{0.27} & 0.31 \\
    \method~(ours) &  \cellcolor{sdf} \textbf{0.13} &   \cellcolor{sdf}  \textbf{0.56} &  0.11 &   \cellcolor{sdf} \textbf{0.71} & 0.11 & 0.66 & 0.13 & \cellcolor{sdf}  \textbf{0.72} & &  \cellcolor{sdf}\textbf{0.49}&   \cellcolor{sdf} \textbf{0.23} &\cellcolor{sdf}  \textbf{0.28} &   \cellcolor{sdf} \textbf{0.46} & \cellcolor{sdf} \textbf{0.33} &  \cellcolor{sdf} \textbf{0.29} & 0.35 &   \cellcolor{sdf} \textbf{0.34}\\

    \cmidrule[\lightrulewidth]{1-9}
    \cmidrule[\lightrulewidth]{11-18}

    COLMAP~\cite{schoenberger2016mvs} & \cellcolor{mvs} \textbf{0.11} & 0.70 &\cellcolor{mvs} \textbf{0.09}& \cellcolor{mvs} \textbf{0.81} & \cellcolor{mvs} \textbf{0.08 }& 0.78 &\cellcolor{mvs} \textbf{0.08} & 0.74 & & 0.53 & 0.41 & \cellcolor{mvs} \textbf{0.85} & 0.52 &   0.65 & 0.53& \cellcolor{mvs}\textbf{0.26} & 0.63 \\
    OpenMVS~\cite{openmvs2020} & \cellcolor{mvs} \textbf{0.11} & \cellcolor{mvs}\textbf{0.77} & \cellcolor{mvs}\textbf{0.08}& \cellcolor{mvs} \textbf{0.80 }&  \cellcolor{mvs} \textbf{0.08}& \cellcolor{mvs} \textbf{0.82} & 0.10 &  \cellcolor{mvs} \textbf{0.80} & &  0.62 & \cellcolor{mvs} \textbf{0.56} & 1.0 & 0.53& 0.73 & 0.58&  0.31 & \cellcolor{mvs}\textbf{0.64} \\
    GOF~\cite{Yu2024GOF} - sparse &  -- & -- &  0.29 & 0.53&    0.23 &  0.63&  -- &  -- & &  \cellcolor{mvs} \textbf{0.45} &  0.51 & 0.90 &  0.54 &  0.42&  
    0.59&   0.51 &  0.45 \\

      GOF~\cite{Yu2024GOF} - dense &  0.17 & 0.71 &  0.16 &  0.72 &    0.20 &   0.74 &   0.11 & \cellcolor{mvs} \textbf{0.80} & &  0.51 &  0.48 & 0.86 & \cellcolor{mvs} \textbf{0.62} &  \cellcolor{mvs} \textbf{0.41}& \cellcolor{mvs} \textbf{0.67}&   0.30 & 0.55 \\
    \cmidrule[\lightrulewidth]{1-9}
    \cmidrule[\lightrulewidth]{11-18}

    \\
    & \multicolumn{8}{c}{Waymo~\cite{Waymo}}  
    & & \multicolumn{8}{c}{nuScenes~\cite{nuscenes}}\\
    \cmidrule[\heavyrulewidth]{2-9}
    \cmidrule[\heavyrulewidth]{11-18}
    & \multicolumn{2}{c}{Seq. 10061} & \multicolumn{2}{c}{Seq. 13196} & \multicolumn{2}{c}{Seq. 14869} & \multicolumn{2}{c}{Seq. 102751} & &    
    \multicolumn{2}{c}{Seq. 0034} & \multicolumn{2}{c}{Seq. 0071} & \multicolumn{2}{c}{Seq. 0664} & \multicolumn{2}{c}{Seq. 0916}\\
    \cmidrule(lr){2-3} \cmidrule(lr){4-5} \cmidrule(lr){6-7}\cmidrule(lr){8-9}
    \cmidrule(lr){11-12} \cmidrule(lr){13-14} \cmidrule(lr){15-16}\cmidrule(lr){17-18}
    & P$\rightarrow$M & Prec.  & P$\rightarrow$M & Prec. & P$\rightarrow$M & Prec. & P$\rightarrow$M~ & Prec. &
    & P$\rightarrow$M & Prec. & P$\rightarrow$M & Prec. & P$\rightarrow$M & Prec. & P$\rightarrow$M & Prec.  \\
    \cmidrule[\lightrulewidth]{1-9}
    \cmidrule[\lightrulewidth]{11-18}

    StreetSurf~\cite{streetsurf} &  0.22 & 0.43 & 0.35 & 0.53 & 0.23 & 0.35 & 0.25 & 0.24 & & 0.57 & 0.29 & 0.78 & 0.47 & 0.67 & 0.50 & 0.65 & 0.28 \\
    \method~(ours) & \cellcolor{sdf} \textbf{ 0.19 }& \cellcolor{sdf}\textbf{ 0.44} &\cellcolor{sdf} \textbf{0.22} &\cellcolor{sdf} \textbf{0.48} &\cellcolor{sdf} \textbf{0.14}& \cellcolor{sdf} \textbf{0.47} & \cellcolor{sdf}\textbf{0.19}&  \cellcolor{sdf}\textbf{0.30} & &\cellcolor{sdf} \textbf{0.40} & \cellcolor{sdf}\textbf{0.20} &\cellcolor{sdf}\textbf{0.22} &\cellcolor{sdf}\textbf{ 0.59} &\cellcolor{sdf}\textbf{0.40} &\cellcolor{sdf} \textbf{0.40}& \cellcolor{sdf} \textbf{0.22} & \cellcolor{sdf} \textbf{0.54}\\

    \cmidrule[\lightrulewidth]{1-9}
    \cmidrule[\lightrulewidth]{11-18}
    COLMAP~\cite{schoenberger2016mvs} &\cellcolor{mvs} \textbf{0.18} & 0.57 & \cellcolor{mvs} \textbf{0.71} & \cellcolor{mvs} \textbf{0.53} & \cellcolor{mvs}\textbf{ 0.14}& \cellcolor{mvs} \textbf{0.67} & \cellcolor{mvs} \textbf{0.19} &0.59 & & \cellcolor{mvs}\textbf{0.25} & 0.51 & \cellcolor{mvs}\textbf{0.77} & 0.67 & \cellcolor{mvs} \textbf{0.65} & 0.62 & \cellcolor{mvs}\textbf{0.70} & 0.68 \\
    OpenMVS~\cite{openmvs2020} &  0.25 & \cellcolor{mvs} \textbf{0.58} &  0.76 & \cellcolor{mvs}\textbf{0.53}&  0.20 & \cellcolor{mvs} \textbf{0.67} & 0.22 & \cellcolor{mvs} \textbf{0.64} & &  0.28 & \cellcolor{mvs} \textbf{0.54} &  0.85 & \cellcolor{mvs}\textbf{0.69} & 0.75&\cellcolor{mvs}  \textbf{
    0.70}&  0.74 & \cellcolor{mvs}\textbf{0.69} \\
    GOF~\cite{Yu2024GOF} - sparse & 1.87 & 0.32 & 2.32 & 0.20 & 1.63 & 0.36 & 1.54 & 0.29 & & 1.55 & 0.07 & 1.72 & 0.16 &1.49 & 0.12 & 1.41 & 0.18 \\

    GOF~\cite{Yu2024GOF} - dense & 1.20 & 0.38 & 1.17 & 0.39 & 1.55 & 0.41 & 2.11 & 0.34 & & 1.02 & 0.12 & 1.55 & 0.23 &1.44 & 0.12 & 1.06 & 0.29 \\
    \cmidrule[\lightrulewidth]{1-9}
    \cmidrule[\lightrulewidth]{11-18}
  \end{tabular}
    }
    \caption{Quantitative results on KITTI-360 \cite{Kitti}, Pandaset \cite{pandaset}, Waymo Open Dataset \cite{Waymo} and nuScenes \cite{nuscenes}. We report the mean Point to Mesh (P$\rightarrow$M) distance in meters $m$, and the percentage of points with a distance to mesh below 0.15$m$ (Prec.). We highlight best performing implicit SDF methods in \colorbox{sdf}{green} and explicit MVS methods in \colorbox{mvs}{blue}. Missing entry (--) designate failure case. GOF - sparse refers to GOF~\cite{Yu2024GOF} gaussians initialized with sparse point cloud. GOF - dense refers to GOF~\cite{Yu2024GOF} gaussians initialized with dense point cloud.
    }
    \vspace{-0.3cm}
    \label{tab:quantitative_results}
\end{table*}

%% file: sec/4_experiments.tex
\section{Experiments}
We evaluate the performance of \method~on multiple driving scenes and provide qualitative and quantitative comparisons compared to relevant methods. 
\subsection{Implementation details.} We use hash encoding to encode the positions~\cite{mueller2022instant} (with the same default parameters as in InstantNGP~\cite{mueller2022instant}), and spherical harmonics to encode the viewing directions. We use 2 layers with 64 hidden units for the MLPs $\mathcal{F}_\Theta^h$ and $\mathcal{F}_\Theta^c$. Pure volumetric stage last for the 100 first steps and the hybrid stage ends at $35\%$ of the total training time. We perform our experiments on a single high-tier GPU using Adam optimizer with a cosine learning rate decay from $10^{-2}$ to $10^{-4}$. For the parameter $s$, we apply a slightly slower weight decay ($10^{-3}$ to $10^{-5}$) to stabilize convergence. We train our model for a total of 14k iterations. Following the literature, we use Marching Cubes~\cite{10.1145/37401.37422} to generate the final mesh that represents the scene. Further details on implementation can be found in the supplementary materials.
\subsection{Datasets.} 
We evaluate our method on four public autonomous driving datasets: KITTI-360~\cite{Kitti}, nuScenes~\cite{nuscenes}, Waymo Open Dataset~\cite{Waymo} and Pandaset~\cite{pandaset}. We conduct our experiments on four static scenes from each one of these datasets. Since our method is designed for reconstructing the static background of urban areas with limited image overlap, we mask dynamic vehicles and pedestrians in the Pandaset sequences during training. Further details on the selected sequences are provided in the supplementary materials. We use poses provided by the datasets except for Waymo where we recompute the vehicle trajectory and sensor calibration with MOISST~\cite{Herau_2023} as the ones provided are not accurate.

\subsection{Baseline.}
We evaluate the state-of-the-art (SoTA) method StreetSurf~\cite{streetsurf} for driving scenes surface reconstruction, using its official implementation with monocular depth supervision and trajectory optimization disabled for a fair comparaison. All experiments were conducted on the same computational setup.
GOF~\cite{Yu2024GOF}, a Gaussian splitting method for landmark surface reconstruction, was also evaluated using the official implementation. We modified it to initialize the Gaussians with a dense point cloud from COLMAP and adjusted the rasterizer for cameras with an off-center focal point. We evaluate both sparse and dense point cloud initialization.
We extensively tested Neuralangelo~\cite{li2023neuralangelo} on the selected driving sequences, with various training strategies and hyperparameters tuning. However we did not obtain satisfactory results: Neuralangelo requires a bounded scene to initialize the SDF in a sphere shape, making it unsuitable for the open, complex nature of driving sequences. For more details on these experiments, we refer readers to our supplementary materials.
To provide a more comprehensive comparison we benchmark \method~mesh reconstruction with meshes obtained by two methods based on classic MVS: Delaunay triangulation implemented in COLMAP~\cite{schoenberger2016mvs} and mesh reconstruction~\cite{Jancosek2014ExploitingVI} from OpenMVS~\cite{openmvs2020}. For both of the MVS meshing baselines, we used the dense point cloud generated by COLMAP as it provides denser results compared to the one obtained with OpenMVS library.

\subsection{Evaluation metrics.}
To evaluate the quality of the reconstructed surfaces, we report two metrics: 

\noindent \textbf{Point to Mesh (P$\rightarrow$M):} the mean distances from the ground truth LiDAR points to the predicted SDF-generated mesh. 

\noindent \textbf{Precision (Prec.):} the percentage of LiDAR points with a distance to the mesh below $0.15$m.

\subsection{Results}

\paragraph{Quantitative analysis.}
\label{sec:quantitative}
We report quantitative results on the 4 datasets in the Tab.~\ref{tab:quantitative_results}. Results show that \method~achieves on par results with StreetSurf on KITTI-360 dataset, with a mean P$\rightarrow$M error of $12$ centimeters.
However, for the remaining datasets, \method~ consistently outperforms StreetSurf by achieving lower P$\rightarrow$M error and higher precision among most of the scenes.
The similar performance on KITTI-360 is due to its topology with mainly straight roads and open scenes which are more suitable to the design choices done by StreetSurf. Conversely, the other datasets contain more challenging scenarios with wide, non-straight roads, and a high level of detail (we refer to the qualitative analysis). \method~demonstrates comparable P$\rightarrow$M error with explicit MVS methods on KITTI-360, which can be attributed to the fact that KITTI-360 is an ideal dataset for multi-view reconstruction thanks to the good quality of the images and the large FOV of the cameras mounted on the vehicle. However, in more complex environments with smaller overlaps between observations, \method~consistently outperforms explicit MVS methods by achieving lower P$\rightarrow$M error. Since they use a dense point clouds as initialization, explicit MVS methods can achieve higher precision by reconstructing finer structures (see Qualitative analysis).\\
The results averaged for the 4 datasets give a mean P$\rightarrow$M of \textbf{0.24m} for \method~, two times smaller compared to 0.47m for StreetSurf and a mean precision of \textbf{0.46\%} for \method~ and 0.42\% for StreetSurf (see supplementary materials for additional quantitative evaluation).\\ 

\noindent\textbf{Efficiency.} We report in Tab.~\ref{table:performances} the computational performance of each method. \method~is more efficient than StreetSurf~\cite{streetsurf}, using a single 3D hierarchical hash grid instead of two grids (3D and 4D), with a third of the parameters and half the GPU memory. GOF, on the other hand, has variable GPU memory and parameters depending on the number of Gaussians per scene. For a single scene, \method~requires half the training time compared to StreetSurf (20 min vs 40 min), while GOF takes at least 1.5 hours, including 30 minutes each for point cloud initialization, training, and mesh extraction.

\input{tables_tex/efficiency_tab.tex}

\begin{figure*}[tb] 

\centering

    \begin{subfigure}{0.495\textwidth}
        \centering
        \scriptsize
        \setlength{\tabcolsep}{0.002\linewidth}
        \renewcommand{\arraystretch}{0.8}
        \begin{tabular}{cccccc}
            &  COLMAP~\cite{schoenberger2016mvs} & OpenMVS~\cite{openmvs2020} &GOF~\cite{Yu2024GOF} &  StreetSurf~\cite{streetsurf} & \method~(ours) \\ 
            \multirow{1}{*}[7.5mm]{\rotatebox[origin=c]{90}{Seq. 30}}  &
            \includegraphics[clip=true, trim={0 0 0 0},width=0.19\columnwidth]{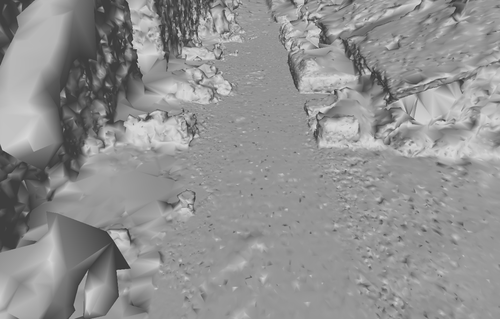} & 
            \includegraphics[clip=true, trim={0 0 0 0},width=0.19\columnwidth]{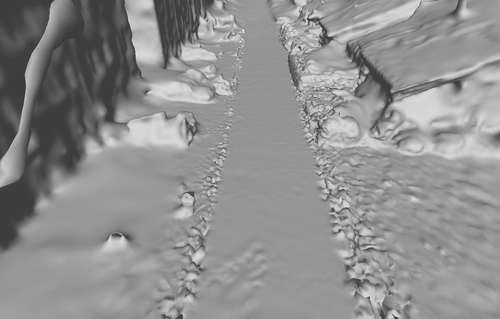} & 
            \includegraphics[clip=true, trim={0 0 0 0},width=0.19\columnwidth]{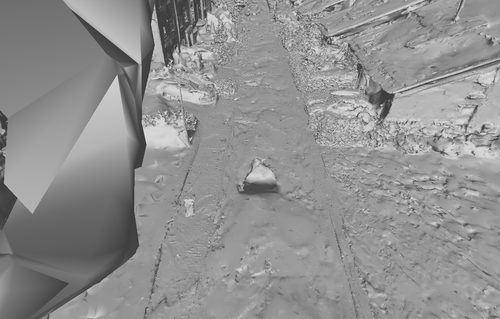} & 
            \includegraphics[clip=true, trim={0 0 0 0},width=0.19\columnwidth]{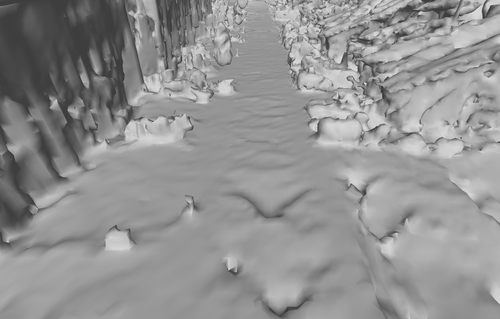} & 
            \includegraphics[clip=true, trim={0 0 0 0},width=0.19\columnwidth]{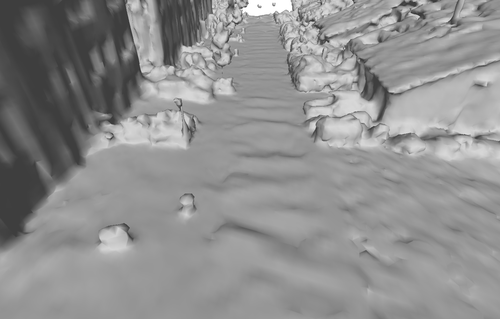} \\
            
            \multirow{1}{*}[7.5mm]{\rotatebox[origin=c]{90}{Seq. 31}}  &
            \includegraphics[clip=true, trim={0 0 0 0},width=0.19\columnwidth]{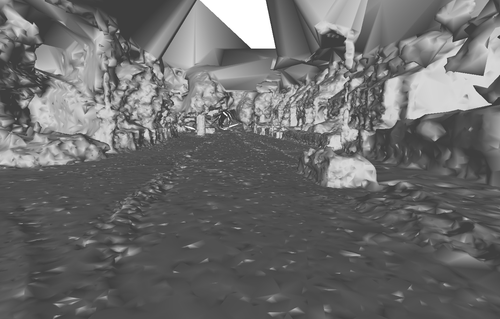} & 
            \includegraphics[clip=true, trim={0 0 0 0},width=0.19\columnwidth]{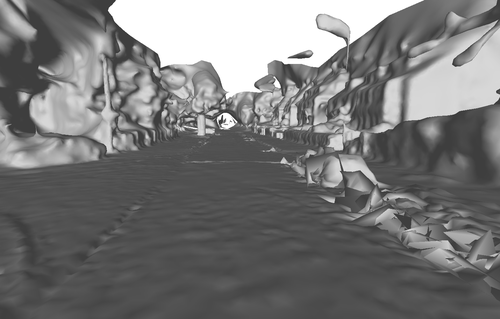} & 
            \includegraphics[clip=true, trim={0 0 0 0},width=0.19\columnwidth]{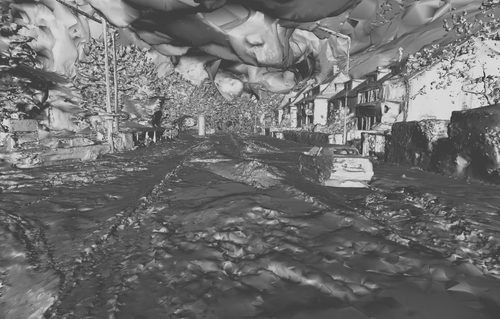} & 
            \includegraphics[clip=true, trim={0 0 0 0},width=0.19\columnwidth]{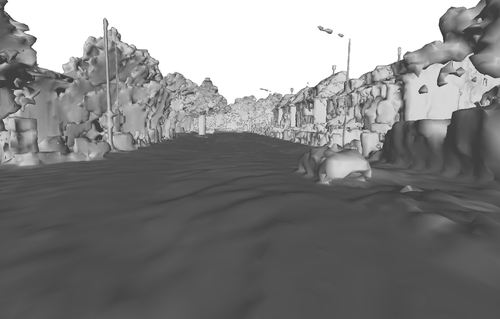} & 
            \includegraphics[clip=true, trim={0 0 0 0},width=0.19\columnwidth]{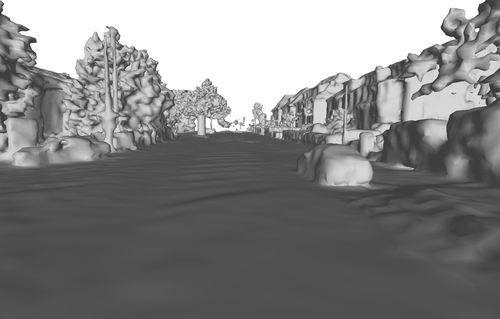} \\
            
        \end{tabular}
        \caption{KITTI-360~\cite{Kitti}} 
        \label{fig:qualitative/kitti}
    \end{subfigure}
    \begin{subfigure}{0.495\textwidth}
        \scriptsize
        \centering
        \setlength{\tabcolsep}{0.002\linewidth}
        \renewcommand{\arraystretch}{0.8}
        \begin{tabular}{cccccc}
            &  COLMAP~\cite{schoenberger2016mvs} & OpenMVS~\cite{openmvs2020} &GOF~\cite{Yu2024GOF} &  StreetSurf~\cite{streetsurf} & \method~(ours) \\ 
            \multirow{1}{*}[7.5mm]{\rotatebox[origin=c]{90}{Seq. 42}}  &
            \includegraphics[clip=true, trim={0 0 0 0},width=0.19\columnwidth]{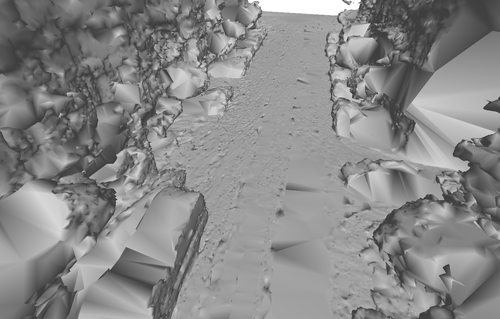} & 
            \includegraphics[clip=true, trim={0 0 0 0},width=0.19\columnwidth]{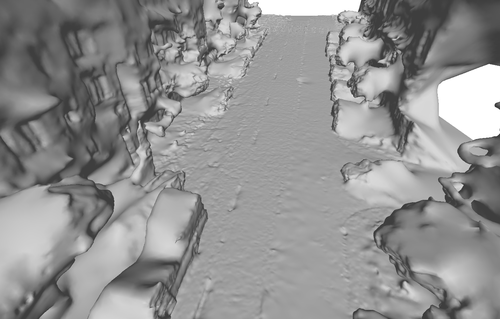} & 
            \includegraphics[clip=true, trim={0 0 0 0},width=0.19\columnwidth]{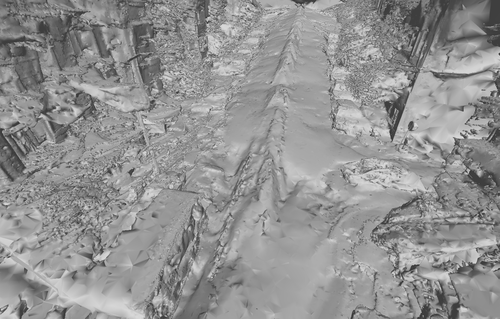} & 
            \includegraphics[clip=true, trim={0 0 0 0},width=0.19\columnwidth]{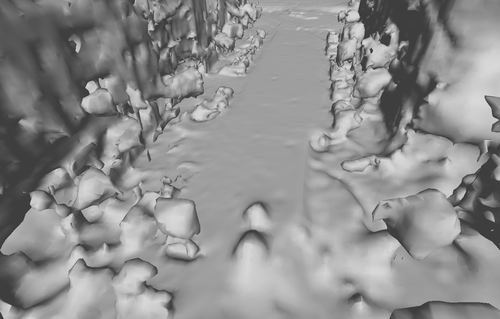} & 
            \includegraphics[clip=true, trim={0 0 0 0},width=0.19\columnwidth]{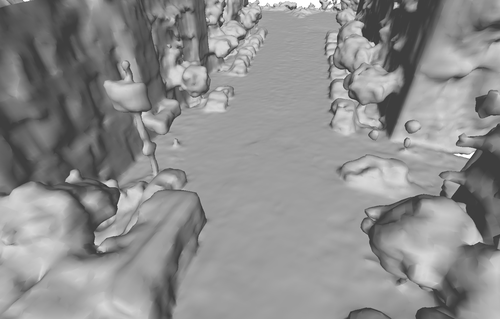} \\
            
             \multirow{1}{*}[7.5mm]{\rotatebox[origin=c]{90}{Seq. 43}}  &
            \includegraphics[clip=true, trim={0 0 0 0},width=0.19\columnwidth]{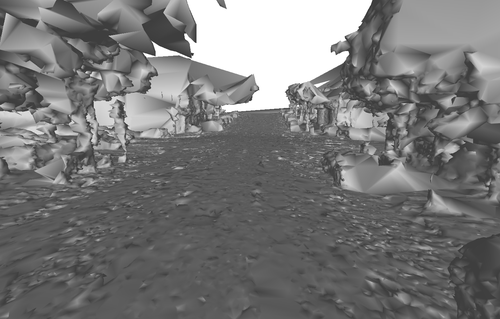} & 
            \includegraphics[clip=true, trim={0 0 0 0},width=0.19\columnwidth]{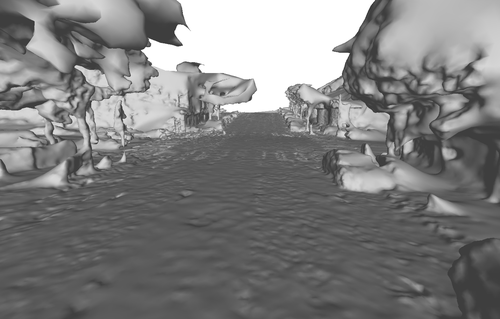} & 
            \includegraphics[clip=true, trim={0 0 0 0},width=0.19\columnwidth]{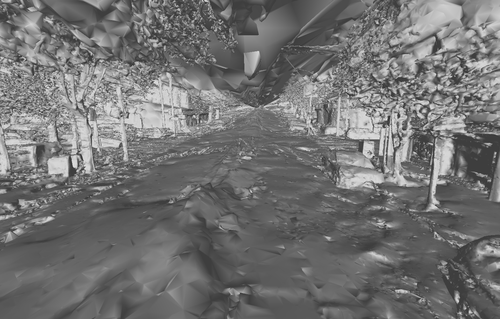} & 
            \includegraphics[clip=true, trim={0 0 0 0},width=0.19\columnwidth]{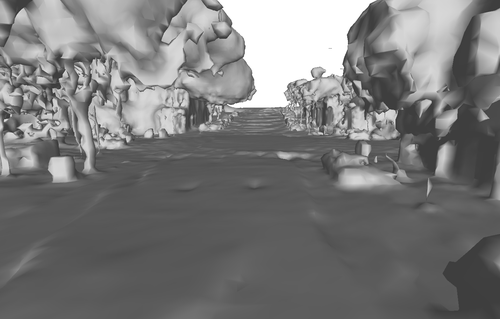} & 
            \includegraphics[clip=true, trim={0 0 0 0},width=0.19\columnwidth]{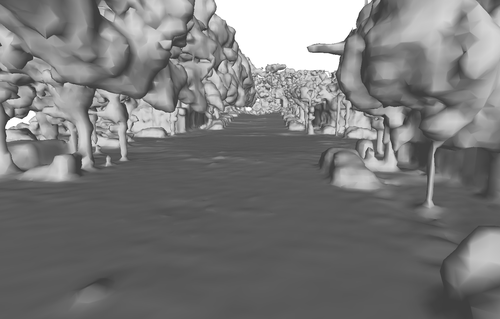} \\
            
        \end{tabular}
        \caption{Pandaset~\cite{pandaset}} 
        \label{fig:qualitative/pandaset}
    \end{subfigure}
    \\
    \vspace{0.2cm}
    \begin{subfigure}{0.495\textwidth}
        \centering
        \scriptsize
        \setlength{\tabcolsep}{0.002\linewidth}
        \renewcommand{\arraystretch}{0.8}
        \begin{tabular}{cccccc}
            \multirow{1}{*}[8.7mm]{\rotatebox[origin=c]{90}{Seq. 0034}}  &
            \includegraphics[clip=true, trim={0 0 0 0},width=0.19\columnwidth]{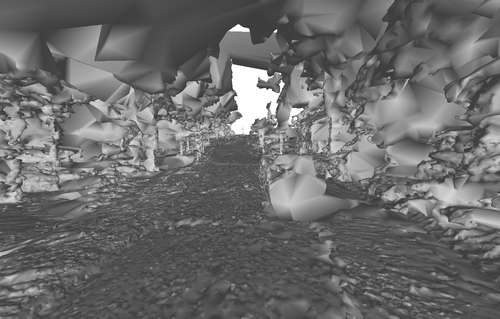} & 
            \includegraphics[clip=true, trim={0 0 0 0},width=0.19\columnwidth]{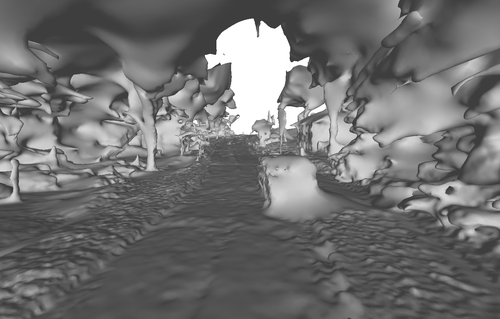} & 
            \includegraphics[clip=true, trim={0 0 0 0},width=0.19\columnwidth]{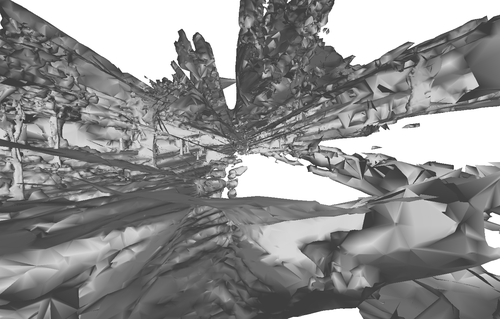} & 
            \includegraphics[clip=true, trim={0 0 0 0},width=0.19\columnwidth]{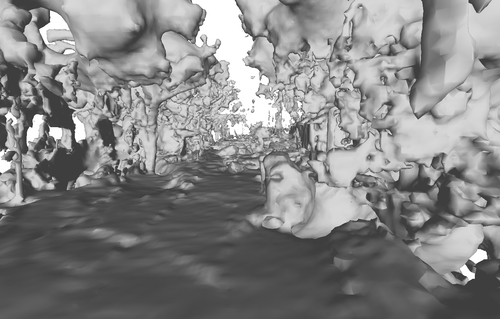} & 
            \includegraphics[clip=true, trim={0 0 0 0},width=0.19\columnwidth]{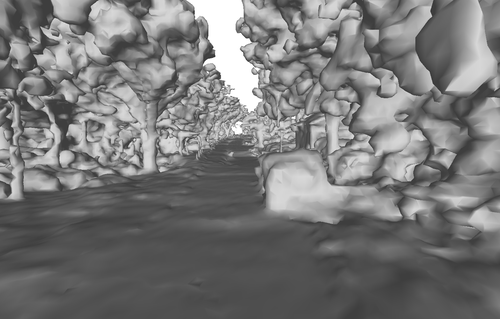} \\
            
            \multirow{1}{*}[8.0mm]{\rotatebox[origin=c]{90}{Seq. 916}}  &
            \includegraphics[clip=true, trim={0 0 0 0},width=0.19\columnwidth]{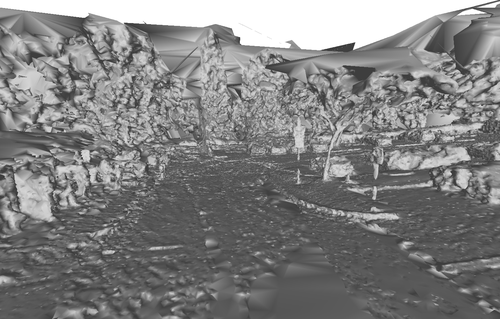} & 
            \includegraphics[clip=true, trim={0 0 0 0},width=0.19\columnwidth]{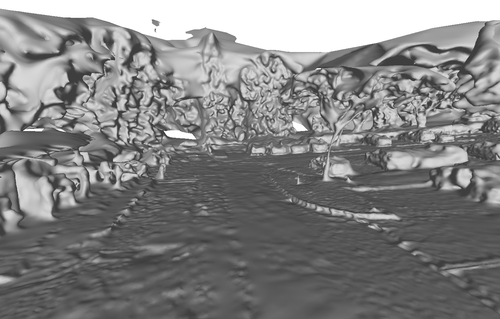} & 
            \includegraphics[clip=true, trim={0 0 0 0},width=0.19\columnwidth]{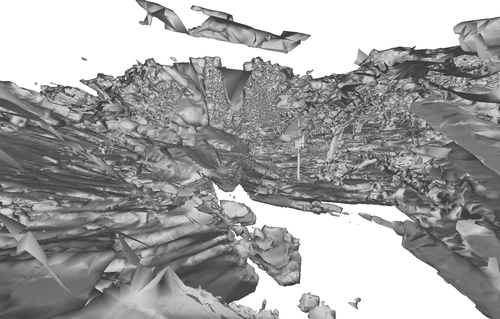} & 
            \includegraphics[clip=true, trim={0 0 0 0},width=0.19\columnwidth]{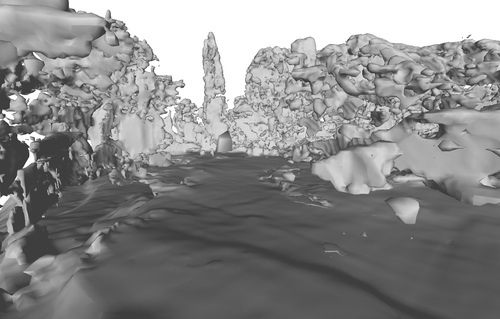} & 
            \includegraphics[clip=true, trim={0 0 0 0},width=0.19\columnwidth]{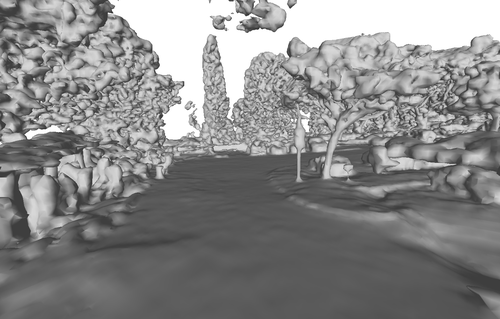} 
        \end{tabular}
        \caption{nuScenes~\cite{nuscenes}} 
        \label{fig:qualitative/nuscenes}
    \end{subfigure}
    \hfill
    \begin{subfigure}{0.495\textwidth}
        \centering
        \scriptsize
        \setlength{\tabcolsep}{0.002\linewidth}
        \renewcommand{\arraystretch}{0.8}
        \begin{tabular}{cccccc}
            \multirow{1}{*}[7.5mm]{\rotatebox[origin=c]{90}{Seq. 10}}  &
            \includegraphics[clip=true, trim={0 0 0 0},width=0.19\columnwidth]{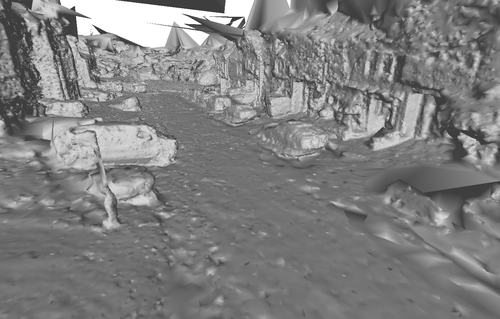} & 
            \includegraphics[clip=true, trim={0 0 0 0},width=0.19\columnwidth]{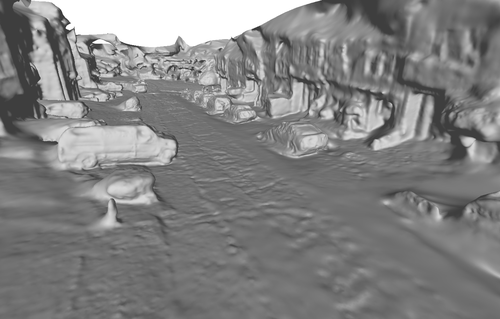} & 
            \includegraphics[clip=true, trim={0 0 0 0},width=0.19\columnwidth]{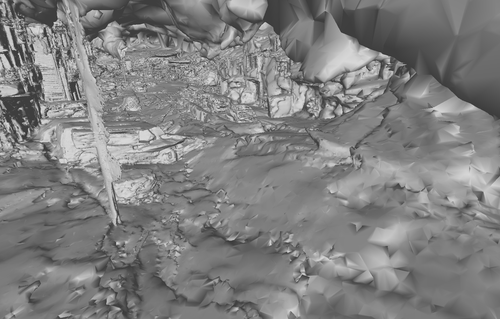} & 
            \includegraphics[clip=true, trim={0 0 0 0},width=0.19\columnwidth]{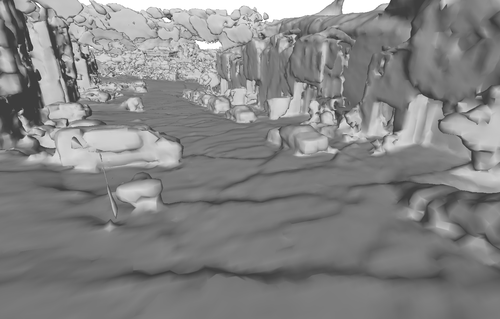} & 
            \includegraphics[clip=true, trim={0 0 0 0},width=0.19\columnwidth]{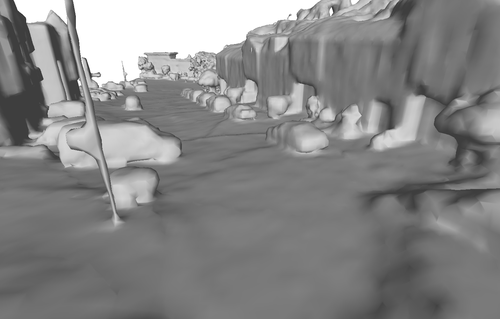} \\
            
            \multirow{1}{*}[8.0mm]{\rotatebox[origin=c]{90}{Seq. 13}}  &
            \includegraphics[clip=true, trim={0 0 0 0},width=0.19\columnwidth]{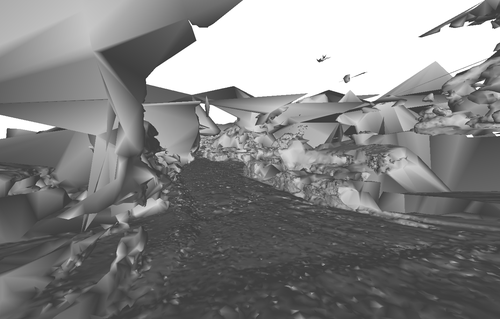} & 
            \includegraphics[clip=true, trim={0 0 0 0},width=0.19\columnwidth]{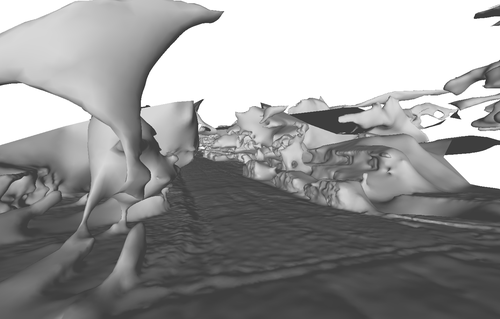} & 
            \includegraphics[clip=true, trim={0 0 0 0},width=0.19\columnwidth]{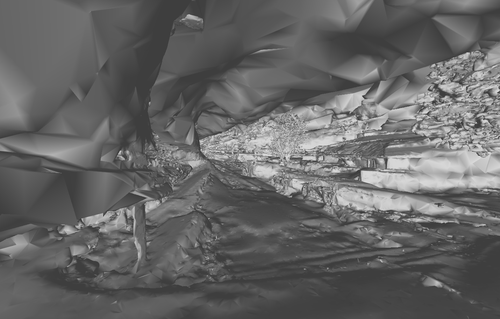} & 
            \includegraphics[clip=true, trim={0 0 0 0},width=0.19\columnwidth]{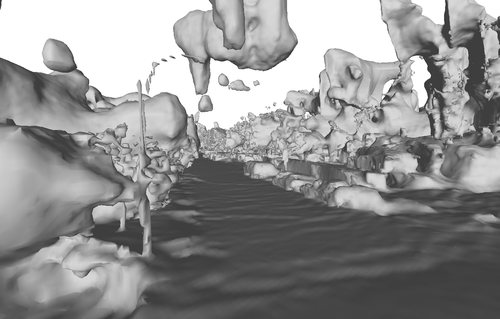} & 
            \includegraphics[clip=true, trim={0 0 0 0},width=0.19\columnwidth]{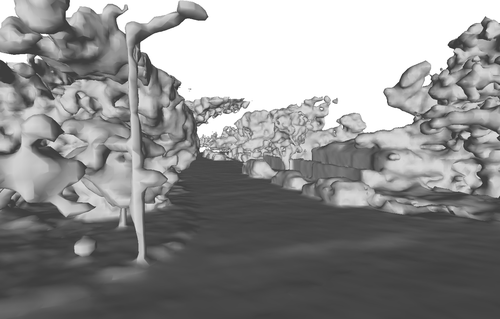}
        \end{tabular}
        \caption{Waymo Open Dataset~\cite{Waymo}} 
        \label{fig:qualitative/waymo}
    \end{subfigure}
  
      \caption{Qualitative experiments results on \protect\subref{fig:qualitative/kitti} KITTI-360, \protect\subref{fig:qualitative/pandaset} Pandaset, \protect\subref{fig:qualitative/nuscenes} nuScenes and \protect\subref{fig:qualitative/waymo} Waymo Open Dataset. 
      We compare our mesh extracted from our SDF to GOF, COLMAP, OpenMVS and StreetSurf meshes.}
   \label{fig:qualitative-pandaset}
\end{figure*}

\vspace{-0.5cm}
\paragraph{Qualitative analysis.}
\label{sec:qualitative}
The quantitative evaluation presented in the previous section (Sec. \ref{sec:quantitative}) is computed from the accumulated ground-truth LiDAR which are sparse and do not cover some regions of the scene. To complete our evaluation, we present qualitative results in Figs. ~\ref{fig:qualitative-pandaset}.
Results demonstrate that \method~reconstructs higher-quality surfaces compared to StreetSurf and explicit MVS methods. Even if finer details can be recovered on some parts of the scene, meshes obtained by explicit MVS methods have noisy normals and are incomplete, which limits their applicability to some tasks such as mesh texturing (see supplementary materials).
We observe that our method recovers more accurate and realistic scene details at many regions such as buildings, cars, poles, and trees.  
Our analysis supports the hypothesis that SteertSurf's disentanglement of the close and far range based on the assumption that driving scenes are long and narrow is not adapted to the majority of driving sequences. 
In addition, StreetSurf's road-surface initialization fails in sequences where the road and sidewalks are not at the same height or at non-flat roads such as downhills (see Fig.~\ref{fig:qualitative-pandaset} -- Seq. 30).

\subsection{Ablation study}

\begin{figure}[tb]
\centering
\begin{subfigure}{0.49\linewidth}
    \centering
    \includegraphics[width=1.0\textwidth]{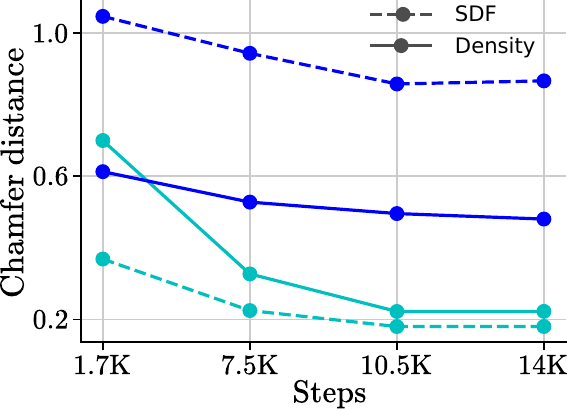} 
    \caption{}
\label{fig:ablations-VolSDF}
\end{subfigure}
\begin{subfigure}{0.49\linewidth}
    \centering
    \includegraphics[width=1.0\textwidth]{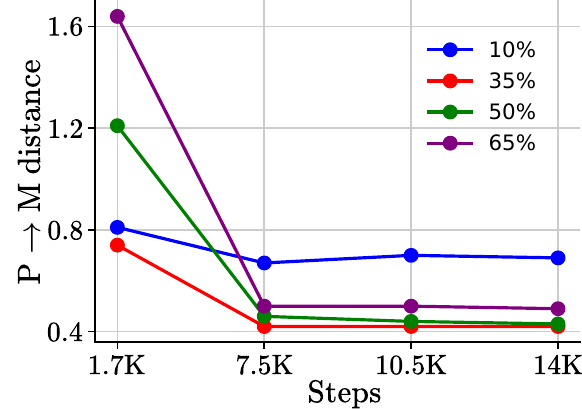} 
    \caption{} 
    \label{fig:ablations-P2M}
\end{subfigure}
\vspace{-2mm}
\caption{Chamfer distance between LiDAR and predicted point cloud of Density field and SDF field at different training steps (left). P$\rightarrow$M for different hybrid stage duration at different training steps for Seq. 23  from Pandaset (right).}
\label{fig:vol_vs_sdf}
\vspace{-5mm}
\end{figure}

\paragraph{Convergence of density field compared to SDF field.}
We report in Fig.~\ref{fig:vol_vs_sdf} the Chamfer Distance between LiDAR and predicted point clouds, for Seq. 31 from KITTI-360 and Seq. 916 from nuScenes, comparing two different NeRF: one solely based on density volumetric field and another one solely based on SDF at various training steps (without normal supervision, as we cannot supervise normal in case of pure density volumetric field). The results support the claim that density field converges faster compared to the SDF field.

\paragraph{Duration of hybrid stage.}
To analyze the effectiveness of our progressive initialization approach, we report the P$\rightarrow$M distance at various iterations with different transition times in Fig.~\ref{fig:vol_vs_sdf}. The results demonstrate that a short hybrid stage cannot be sufficient to initialize the SDF field, while a long hybrid stage contributes to losing the initialization of the SDF field and results in sub-optimal reconstructions.

\paragraph{\method's components.}
We ablate the effect of \method's SDF progressive initialization strategy and show quantitative results in Tab.~\ref{table:ablations}. The results demonstrate that our initialization that transitions from the volumetric to the surface representation is crucial to accurately initialize the SDF field and efficiently reconstruct driving sequences. In addition, we report our method with standard accumulated normal supervision, and without the network regularization loss.

\begin{table}[t]
    \caption{Ablations study: we deactivate some of the key components of \method~in order to measure the impact on the final reconstruction. (a) without progressive initialization (b) with the accumulated normal for $\mathcal{L}_{\hat{N}}$ (c) without $\mathcal{L}_{\text{s}}$.}
    \centering
    \resizebox{1.0\columnwidth}{!}{%
    \begin{tabular}{ccccccccc}
        & \multicolumn{2}{c}{w/o SDF init} & \multicolumn{2}{c}{w/o $\mathcal{L}_{\hat{N}}$} & \multicolumn{2}{c}{w/o $\mathcal{L}_{\text{s}}$} & \multicolumn{2}{c}{\method} \\ 
        \cmidrule[\lightrulewidth]{2-9}
       Seq.  & P$\rightarrow$M & Prec.  & P$\rightarrow$M & Prec. & P$\rightarrow$M & Prec. & P$\rightarrow$M~ & Prec. \\
      \cmidrule(lr){1-1}  \cmidrule(lr){2-3} \cmidrule(lr){4-5} \cmidrule(lr){6-7}\cmidrule(lr){8-9}
         KITTI-360. 31&  0.35 & 0.18  & 0.19 & 0.27 & 0.24 & 0.15 & \textbf{0.15} &\textbf{ 0.71}  \\
         \bottomrule
        Pandaset. 23 & 0.85 & 0.18  & 0.80 & 0.18 & 0.54 & 0.17 &\textbf{ 0.51} & \textbf{0.23} \\
        \bottomrule
         nuScenes. 0916& 0.35 & 0.4 & 0.4 & 0.21 & 0.35 & 0.15 & \textbf{0.25} &\textbf{ 0.54 } \\
         \bottomrule
         Waymo. 10061 & 0.3 & 0.36 & 0.22 & 0.32 & 0.34 & 0.17 & \textbf{0.23} & \textbf{0.43} \\
 
        \bottomrule
        
    \end{tabular}}
    \label{table:ablations}
   \vspace{-5mm} 
\end{table}

%% file: tables_tex/efficiency_tab.tex
\begin{table}[tb]
    
    \centering
    \resizebox{0.8\columnwidth}{!}{%
    \begin{tabular}{lccc}
        &Total train time &GPU memory&Params size   \\
        &(min)&(Gb)&(MiB)\\
        
        \midrule
        GOF training$^\dagger$~\cite{Yu2024GOF} & 30-60 & 8-19 & 100-300   \\
        GOF mesh extraction~\cite{Yu2024GOF} & 30 & 8 & -   \\
        StreetSurf~\cite{streetsurf} & 40 & 19 & 92.59  \\
        \method~(ours) & \textbf{20} & \textbf{8} & \textbf{27} \\
        \midrule
     
        COLMAP~\cite{schoenberger2016mvs} & 30 & $<$1 & -- \\
        OpenMVS~\cite{openmvs2020} & 30  & $<$1 & -- \\
        \bottomrule
        \vspace{-0.5cm}
    \end{tabular}
    }
        \caption{\textbf{Training analysis:} reported using the same GPU, equivalent to an RTX4090. $^\dagger$ Training time excludes COLMAP point cloud initialization.}
    \label{table:performances}
\end{table}

%% file: sec/5_discussion.tex
\section{Discussion}
\noindent\textbf{Applications.}
Thanks to the quality and completeness of \method's reconstructed meshes, we can use modern MVS tools such as OpenMVS~\cite{openmvs2020} to obtain a detailed and colorized representation of large urban scenes, as shown in Fig.~\ref{fig:teaser}.

\noindent\textbf{Limitations.}
Although our method can achieve highly detailed surface reconstruction in driving scenarios, it fails to reconstruct fine details in very wide and open sequences. We recommend consulting the supplementary materials for the failure cases of \method.

%% file: sec/6_conclusion.tex
\section{Conclusion}
In this work we presented \textbf{\method}, an implicit hybrid approach to efficiently reconstruct large urban scenes. Our method's hybrid architecture models the volumetric density and signed distance in two separate fields. We introduced a novel hybrid volume rendering strategy to progressively transition from volumetric representation to a signed distance representation. The quantitative and qualitative evaluations that we performed show that \method~achieves more precise reconstruction than state-of-the-art methods while being significantly faster and is better suited for a large variation of driving scenarios.

%% file: sec/X_suppl.tex
\clearpage
\setcounter{page}{1}
\maketitlesupplementary

In this supplementary material, we provide additional implementation details, experiments with Neuralangelo~\cite{li2023neuralangelo} and additional quantitative and qualitative results. Furthermore, an ablation study on our samples attribution strategy, applications, and failure cases of ViiNeuS are also provided. 

\section{ SDF-gradient normalization} 
    As explained in Sec.~\textcolor{red}{3.3} of the main paper, volumetric representations based on density estimation tend to quickly approximate saturated alpha values ($\alpha_i^v$ being either $0$ or $1$), while alpha values computed from the SDF converge slower. Because $\alpha_i^v$ and $\alpha_i^f$ are composed jointly during the hybrid stage, we noticed that the surface representation compensates for this gap by predicting large SDF gradients so that $\alpha_i^f$ aligns with $\alpha_i^v$. Indeed, from Eq.~\textcolor{red}{3} of the main paper, a solution to saturate alpha towards either 1 or 0, is to predict  $f(p_{i+1}) \ll f(p_i)$ or $f(p_{i+1}) \gg f(p_i)$, respectively. During the forward pass, $f(p_i)$ and $f(p_{i+1})$ are not directly predicted by the model but rather computed using:
    \begin{equation}
        \begin{split}
        f(p_i) & = f(x_i) +  \text{Relu}(-\cos({\theta})) \times \frac{\delta_i}{2}, \\
        f(p_{i+1}) & = f(x_i) - \text{Relu}(-\cos({\theta})) \times \frac{\delta_i}{2},
        \end{split}
    \end{equation}
    \noindent with $\theta$ being the angle between the  direction of $\nabla f(x_i)$ and the ray direction $d$. Practically, we do not formally compute $\theta$ but we rather approximate its cosine with $\cos({\theta}) = \nabla f(x_i) \cdot d$, \textbf{assuming both vectors are unit norm}. By predicting $\left \| \nabla f(x_i) \right \|_2 \gg 1$, $\alpha_i^f$ can be easily saturated and follow the distribution of $\alpha_i^v$ without learning a proper signed distance field.
    To address this, we simply normalize the SDF gradient before the cosine computation to prevent the gradient from compensating the alpha distribution difference. It is important to notice that even if the eikonal loss used for training (see main paper) is supposed to encourage the network to model a signed distance function with spatial derivative of unitary norm, we found that numerically normalizing the gradient during our hybrid stage is essential to avoid divergence in early training iterations.

\begin{figure}[tb] 

\centering
    \begin{subfigure}{0.495\textwidth}
        \centering
        \scriptsize
        \setlength{\tabcolsep}{0.002\linewidth}
        \renewcommand{\arraystretch}{0.8}
        \begin{tabular}{cc}         
             Neuralangelo~\cite{li2023neuralangelo} & ViiNeuS~(ours) \\ 
            \includegraphics[clip=true, trim={0 0 0 0},width=0.49\columnwidth]{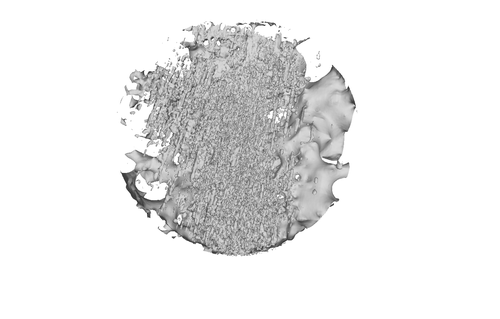} & 
            \includegraphics[clip=true, trim={0 0 0 0},width=0.49\columnwidth]{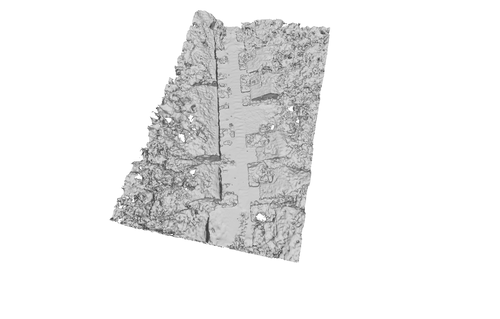} \\
            \includegraphics[clip=true, trim={0 0 0 0},width=0.49\columnwidth]{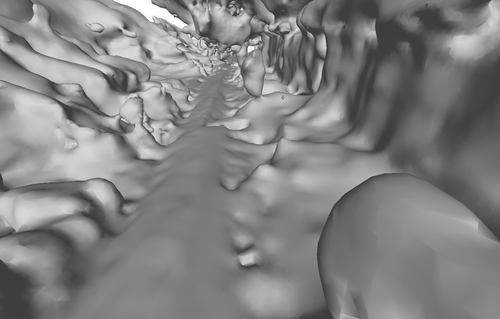} & 
            \includegraphics[clip=true, trim={0 0 0 0},width=0.49\columnwidth]{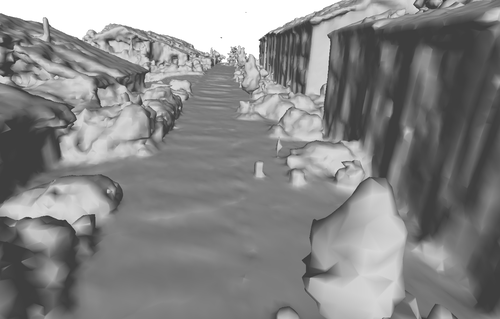} 
        \end{tabular}
        \caption{Seq 30. from KITTI-360~\cite{Kitti}} 
        \label{fig: qualitative/kitti30}
        \vspace{10pt}
    \end{subfigure}
    
    \begin{subfigure}{0.495\textwidth}
        \centering
        \scriptsize
        \setlength{\tabcolsep}{0.002\linewidth}
        \renewcommand{\arraystretch}{0.8}
        \begin{tabular}{cc}
             Neuralangelo~\cite{li2023neuralangelo} & ViiNeuS~(ours) \\          
            \includegraphics[clip=true, trim={0 0 0 0},width=0.49\columnwidth]{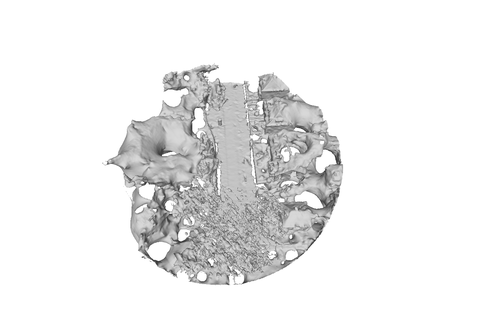} & 
            \includegraphics[clip=true, trim={0 0 0 0},width=0.49\columnwidth]{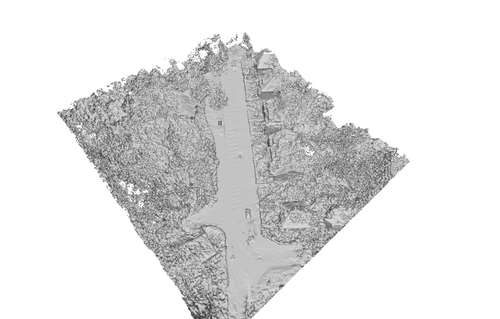} \\
            \includegraphics[clip=true, trim={0 0 0 0},width=0.49\columnwidth]{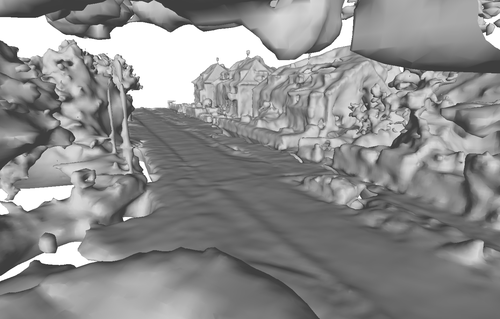} & 
            \includegraphics[clip=true, trim={0 0 0 0},width=0.49\columnwidth]{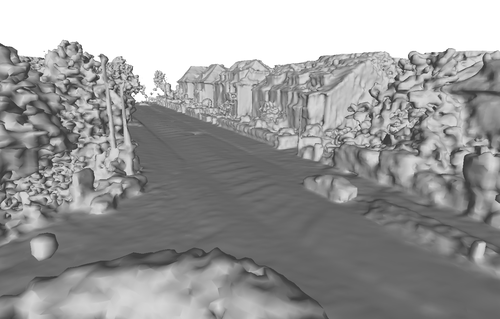} 
        \end{tabular}
        \caption{Seq. 31 from KITTI-360~\cite{Kitti}} 
        \label{fig: qualitative/kitti31}
        \vspace{10pt}
    \end{subfigure}    
     
    \begin{subfigure}{0.495\textwidth}
        \centering
        \scriptsize
        \setlength{\tabcolsep}{0.002\linewidth}
        \renewcommand{\arraystretch}{0.8}
        \begin{tabular}{cc}
             Full seq.  & Segmented sequences \\ 
            \includegraphics[clip=true, trim={0 0 0 0},width=0.49\columnwidth]{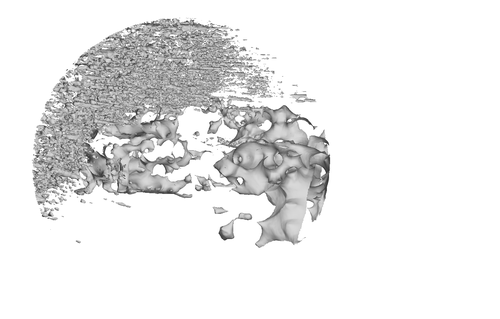} & 
            \includegraphics[clip=true, trim={0 0 0 0},width=0.49\columnwidth]{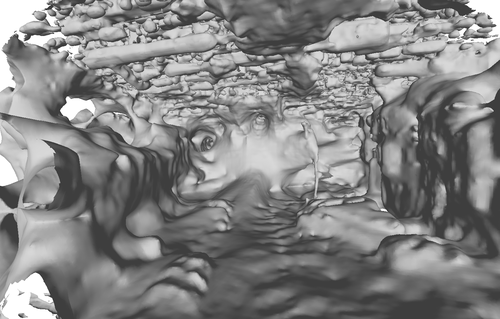} 
        \end{tabular}
        \caption{Failure of Neuralangelo on Seq 10061. from Waymo Open Dataset~\cite{Waymo}} 
        \label{fig: qualitative/waymofail}
    \end{subfigure}
      \caption{Qualitative experiments results on Seq. 30 \protect\subref{fig: qualitative/kitti30} and Seq. 31  \protect\subref{fig: qualitative/kitti31} from KITTI-360s~\cite{Kitti}. Failure on Seq. 10061  \protect\subref{fig: qualitative/waymofail} Waymo Open Dataset~\cite{Waymo}. We compare our generated SDF mesh to Neuralangelo~\cite{li2023neuralangelo} mesh for KITTI-360. We report the experiments on the full and segmented sequence for Waymo.}
  
\end{figure}

\section{SDF Field Initialization: Neuralangelo}

As detailed in the main paper, Neuralangelo~\cite{li2023neuralangelo} is designed for landmark reconstruction, relying on many overlapping images and a bounded region of interest to initialize the SDF with a spherical shape. In our initial tests, when applying Neuralangelo (using the official authors codebase~\footnote{https://github.com/NVlabs/neuralangelo}) to outdoor driving scenarios using the default settings for outdoor scenes, we observed that the initial part of the scene remained noisy and retained a spherical shape. Only the final part of the scene, which was consistently visible across all images in the sequence, was accurately reconstructed (see Fig.~\ref{fig: qualitative/kitti30}).
While Neuralangelo managed to reconstruct some parts of simpler scenes, such as in Seq. 31 from KITTI-360 (Fig.\ref{fig: qualitative/kitti31}), it failed completely in more complex cases, particularly in sequences that are long, wide, or contain challenging structure like downhills (e.g., Seq. 30 in Fig.\ref{fig: qualitative/kitti30}). Moreover, Neuralangelo was unable to reconstruct any part of the Waymo dataset, as shown in Fig.~\ref{fig: qualitative/waymofail}.
Given these outcomes, we conducted extensive experiments with Neuralangelo using various training strategies:

\begin{itemize}

\item Constraining the spherical shape: since Neuralangelo relies on a spherical initialization where everything outside the sphere is treated as background, we attempted to center the sphere on a smaller region, with a smaller radius. However, this approach resulted in very noisy reconstructions, likely due to insufficient overlapping images in that part of the scene (inside the sphere).

\item Segmenting the scene into multiple parts: acknowledging that driving sequences are typically long and wide, we divided the scenes into smaller segments to better fit the spherical initialization. This strategy, however, produced unsatisfactory results. Neuralangelo requires a large number of images (typically 300 for an average Tanks and Temples scene covering the same region of interest) while KITTI-360 sequences contain around 200 images (approximately 50 per camera) and after segmentation, each part of the scene had roughly 100 images. The method could only reconstruct the final segment of the scene that was visible in all images (see Fig.~\ref{fig: qualitative/waymofail}).

\item Doubling the number of images: we further experimented with doubling the number of images used in the reconstruction. Unfortunately, this did not help the method to converge to a satisfactory reconstruction. 
\end{itemize}
Despite various training strategies, Neuralangelo failed to produce reliable reconstructions for complex driving sequences. It requires a large number of overlapping images and a tightly bounded scene, making it unsuitable for unbounded driving sequences. Additionally, Neuralangelo's training time is prohibitively long, taking up to 24 hours per scene. This, combined with its poor performance on driving sequences, makes it impractical for large-scale driving scene reconstruction tasks. 

\section{Additional implementation details}
We use the poses provided by the datasets, except for Waymo, where we recompute the vehicle trajectory and sensor calibration with MOISST~\cite{Herau_2023} due to inaccuracies in the provided data.
The overall loss we use to optimize ViiNeuS is defined as follows: 
\begin{equation}
    \mathcal{L} =  \mathcal{L}_{\text{rgb}} + \lambda_1 \mathcal{L}_{\text{dssim}} + \lambda_2 \mathcal{L}_{\hat{N}} + \lambda_3 \mathcal{L}_{\text{eik}} + \lambda_4 \mathcal{L}_{\text{sky}}.
\end{equation}
We set $\lambda_1$, and $\lambda_4$ to 0.1 and 0.01, respectively. We fix $\lambda_2$ to 0.05 for planar classes and 0.01 for non planar classes. We set $\lambda_3$ to 0.01 in the first training iterations, then we adjust it to 0.1 in the last iterations.\\
We report in table \ref{tab:hashgrid} the hash grid encoding parameters from Instant-NGP \cite{mueller2022instant}. 
We summarize the split of KITTI-360~\cite{Kitti} sequences used for our evaluations in table \ref{tab:kitti}. We use all four cameras for KITTI-360~\cite{Kitti}, and the three front cameras for Pandaset~\cite{pandaset}, nuScenes~\cite{nuscenes} and Waymo Open Dataset~\cite{Waymo}. We sample one image out of two for KITTI-360, and one image out of 8 for the other datasets. 

\begin{table}[]
    \centering
    \caption{Hash grid encoding parameters}
    \begin{tabular}{lcc}
         Parameter & Value \\
         \midrule
         Table size &  $2^{19}$\\
         Finest resolution  & 2048 \\
         Coarsest resolution & 16 \\ 
         Number of level & 16\\ 
        \bottomrule
    \end{tabular}

    \label{tab:hashgrid}
\end{table}

\begin{table}[]
    \centering
    \caption{Selected KITTI-360 sequences}
    \begin{tabular}{lcccccccc}
         Seq. & KITTI Sync. &  Start & End  & \# frames per cam. \\
         \midrule
         30 &  0004 & 1728 & 1822 & 48\\
         31 & 0009 & 2890 & 2996 & 54\\
         35 & 0009 & 980 & 1092 & 57\\ 
        36 & 0010 & 112 & 166 & 28\\ 
        \bottomrule
    \end{tabular}
    
    \label{tab:kitti}
\end{table}

\section{Additional results}

\subsection{Additional quantitative results}

\input{figures_tex/error_curves/error_curves.tex}

We report in Fig~\ref{fig:distance-analysis}  mean cumulative delta error and standard deviation computed across the four sequences from each dataset for both SDF methods. 
As it can be observed, our method's cumulative errors are consistently lower in all datasets for distances below 40$cm$. We additionally show the standard deviation of the error computed along all sequences and notice that our errors remain consistent across the different scenes in contrast to StreetSurf.

\subsection{Additional qualitative results}

\begin{figure*}[h] 

\centering

    \begin{subfigure}{\textwidth}
        \centering
        \scriptsize
        \setlength{\tabcolsep}{0.002\linewidth}
        \renewcommand{\arraystretch}{0.8}
        \begin{tabular}{cccccc}
            &  COLMAP~\cite{schoenberger2016mvs} & OpenMVS~\cite{openmvs2020} &  GOF~\cite{Yu2024GOF} &StreetSurf~\cite{streetsurf} & ViiNeuS~(ours) \\ 

            \multirow{1}{*}[12.5mm]{\rotatebox[origin=c]{90}{Seq. 35}}  &
            \includegraphics[clip=true, trim={0 0 0 0},width=0.19\textwidth]{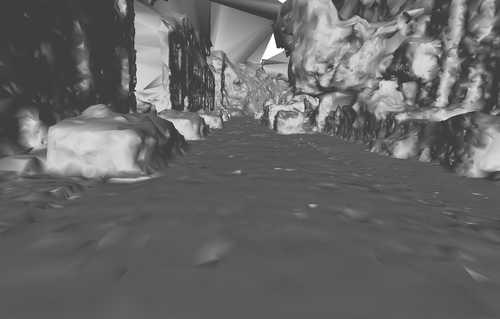} & 
            \includegraphics[clip=true, trim={0 0 0 0},width=0.19\textwidth]{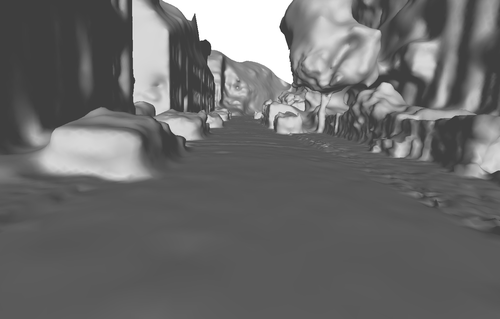} & 
            \includegraphics[clip=true, trim={0 0 0 0},width=0.19\textwidth]{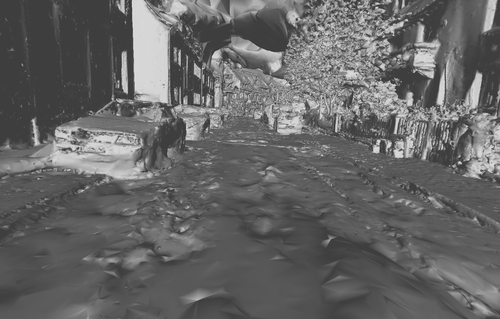} & 
            \includegraphics[clip=true, trim={0 0 0 0},width=0.19\textwidth]{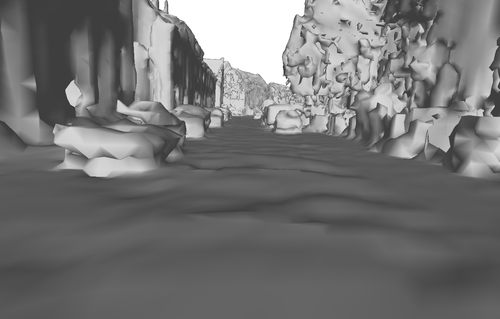} & 
            \includegraphics[clip=true, trim={0 0 0 0},width=0.19\textwidth]{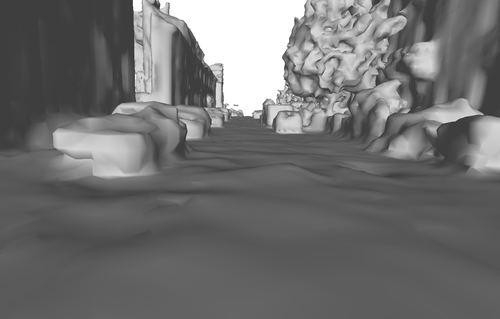} \\   
            \multirow{1}{*}[12.5mm]{\rotatebox[origin=c]{90}{Seq. 36}}  &
            \includegraphics[clip=true, trim={0 0 0 0},width=0.19\textwidth]{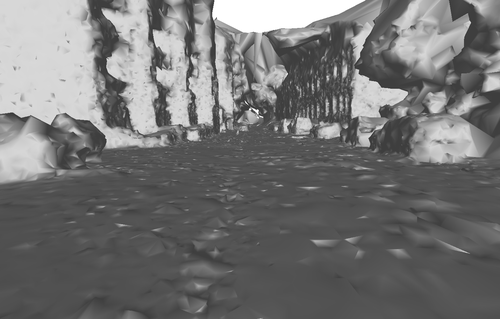} & 
            \includegraphics[clip=true, trim={0 0 0 0},width=0.19\textwidth]{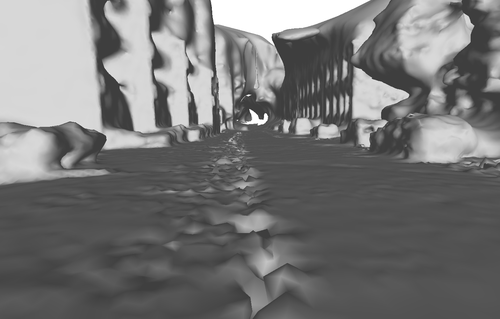} & 
            \includegraphics[clip=true, trim={0 0 0 0},width=0.19\textwidth]{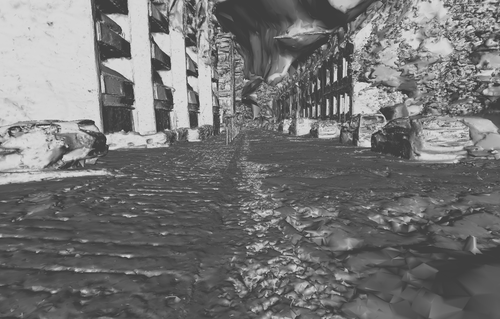} & 
            \includegraphics[clip=true, trim={0 0 0 0},width=0.19\textwidth]{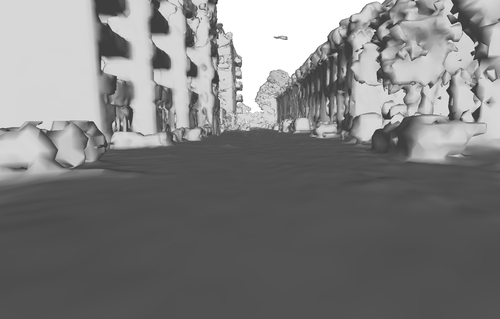} & 
            \includegraphics[clip=true, trim={0 0 0 0},width=0.19\textwidth]{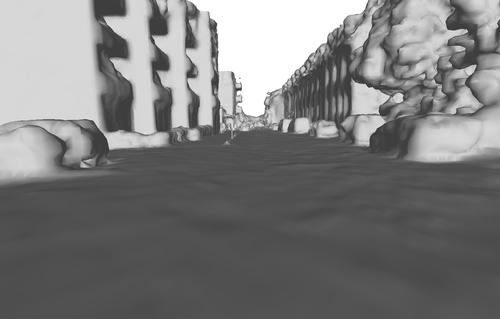} \\    
        \end{tabular}
        \caption{KITTI-360~\cite{Kitti}} 
        \label{fig:qualitative/kitti}
    \end{subfigure}
    \begin{subfigure}{\textwidth}
        \scriptsize
        \centering
        \setlength{\tabcolsep}{0.002\linewidth}
        \renewcommand{\arraystretch}{0.8}
        \begin{tabular}{cccccc}
            \toprule
        
            &  COLMAP~\cite{schoenberger2016mvs} & OpenMVS~\cite{openmvs2020} &  GOF~\cite{Yu2024GOF} &StreetSurf~\cite{streetsurf} & ViiNeuS~(ours) \\

            \multirow{1}{*}[12.5mm]{\rotatebox[origin=c]{90}{Seq. 23}}  &
            \includegraphics[clip=true, trim={0 0 0 0},width=0.19\textwidth]{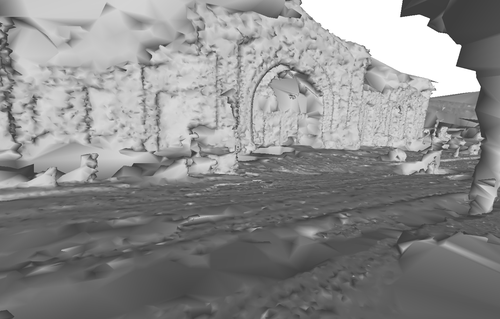} & 
            \includegraphics[clip=true, trim={0 0 0 0},width=0.19\textwidth]{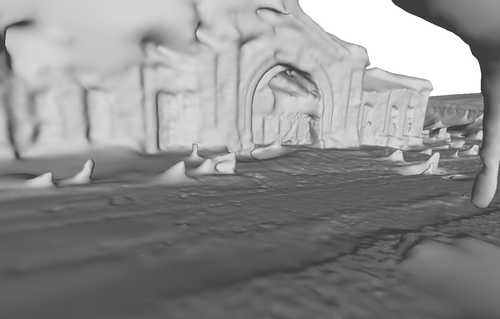} & 
            \includegraphics[clip=true, trim={0 0 0 0},width=0.19\textwidth]{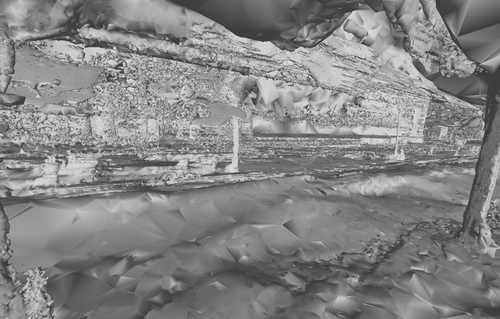} & 
            \includegraphics[clip=true, trim={0 0 0 0},width=0.19\textwidth]{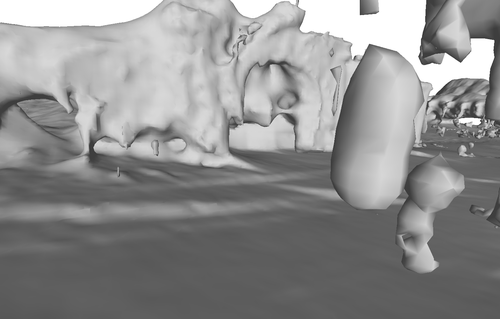} & 
            \includegraphics[clip=true, trim={0 0 0 0},width=0.19\textwidth]{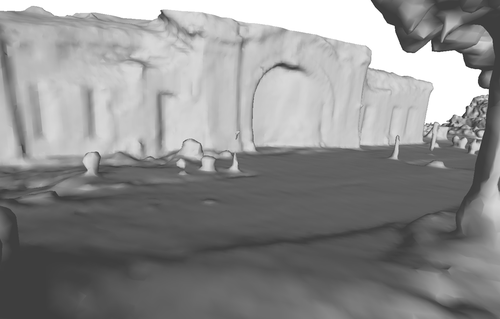} \\

            \multirow{1}{*}[12.5mm]{\rotatebox[origin=c]{90}{Seq. 37}}  &
            \includegraphics[clip=true, trim={44 0 0 0},width=0.19\textwidth]{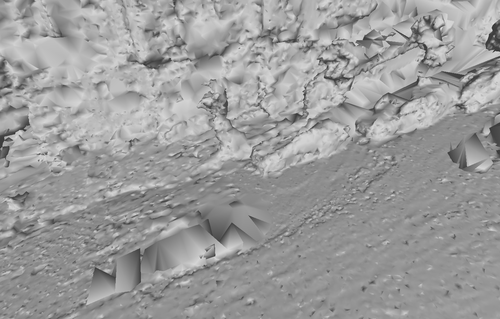} & 
            \includegraphics[clip=true, trim={0 0 0 0},width=0.19\textwidth]{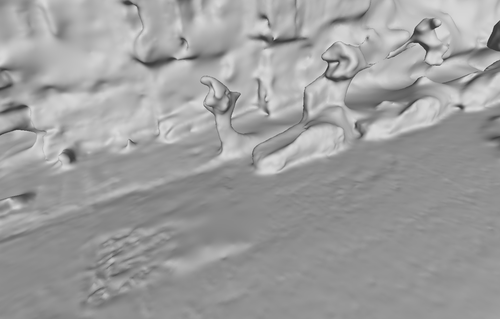} & 
            \includegraphics[clip=true, trim={0 0 0 0},width=0.19\textwidth]{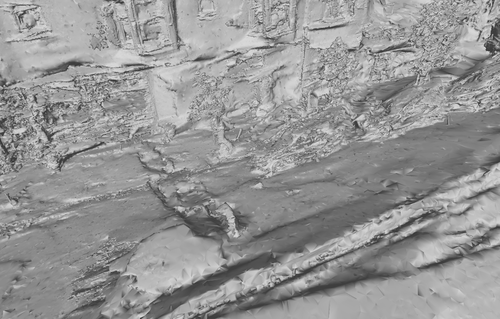} & 
            \includegraphics[clip=true, trim={0 0 0 0},width=0.19\textwidth]{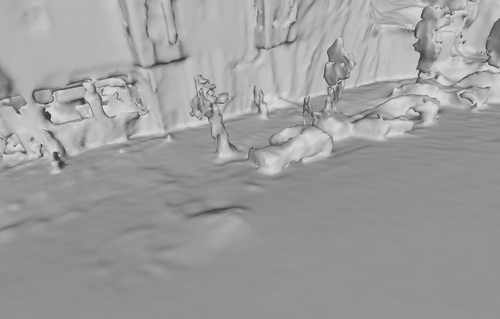} & 
            \includegraphics[clip=true, trim={0 0 0 0},width=0.19\textwidth]{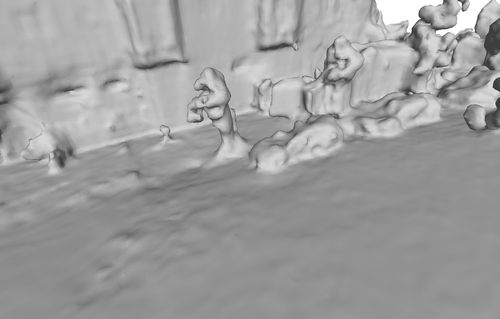} \\
        \end{tabular}
        \caption{Pandaset~\cite{pandaset}} 
        \label{fig:qualitative/pandaset}
    \end{subfigure}
    \\
    \vspace{0.2cm}
    \begin{subfigure}{\textwidth}
        \centering
        \scriptsize
        \setlength{\tabcolsep}{0.002\linewidth}
        \renewcommand{\arraystretch}{0.8}
        \begin{tabular}{cccccc}
            \toprule
        
            &  COLMAP~\cite{schoenberger2016mvs} & OpenMVS~\cite{openmvs2020} &  GOF~\cite{Yu2024GOF} &StreetSurf~\cite{streetsurf} & ViiNeuS~(ours) \\ 
            \multirow{1}{*}[12.5mm]{\rotatebox[origin=c]{90}{Seq. 71}}  &
            \includegraphics[clip=true, trim={0 0 0 0},width=0.19\textwidth]{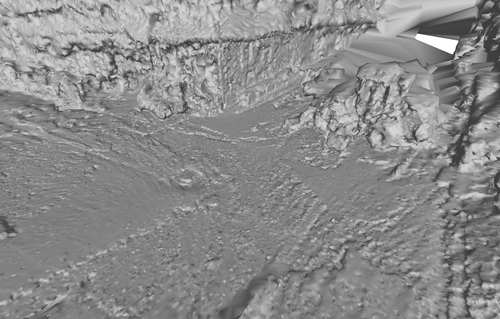} & 
            \includegraphics[clip=true, trim={0 0 0 0},width=0.19\textwidth]{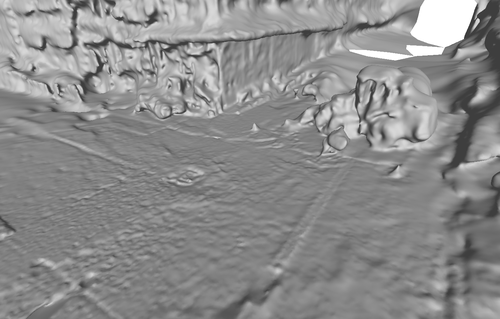} & 
            \includegraphics[clip=true, trim={0 0 0 0},width=0.19\textwidth]{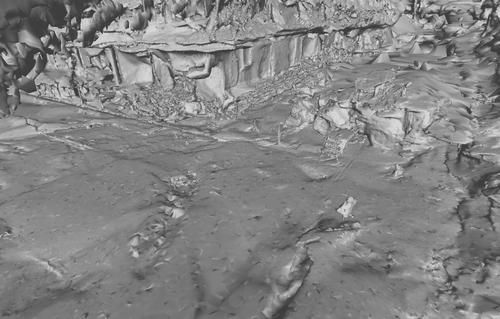} & 
            \includegraphics[clip=true, trim={0 0 0 0},width=0.19\textwidth]{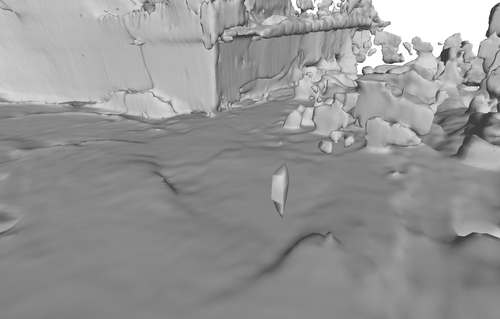} & 
            \includegraphics[clip=true, trim={0 0 0 0},width=0.19\textwidth]{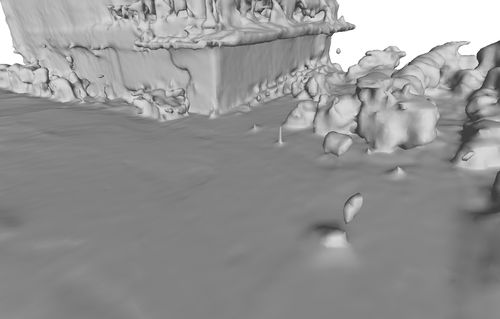} \\
            
            \multirow{1}{*}[12.5mm]{\rotatebox[origin=c]{90}{Seq. 664}}  &
            \includegraphics[clip=true, trim={36 0 0 0},width=0.19\textwidth]{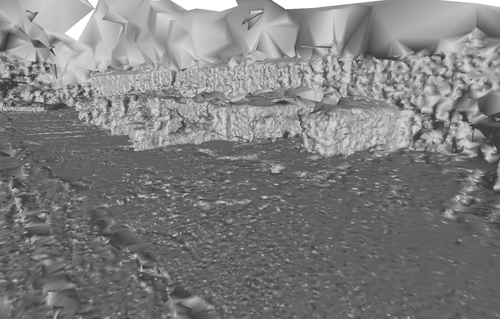} & 
            \includegraphics[clip=true, trim={0 0 0 0},width=0.19\textwidth]{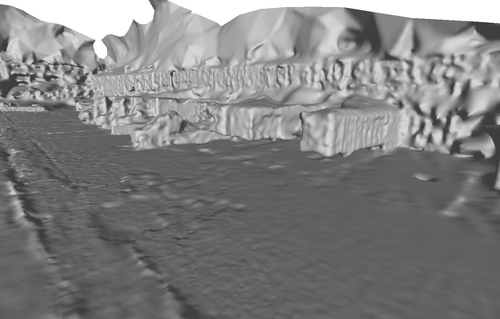} & 
            \includegraphics[clip=true, trim={0 0 0 0},width=0.19\textwidth]{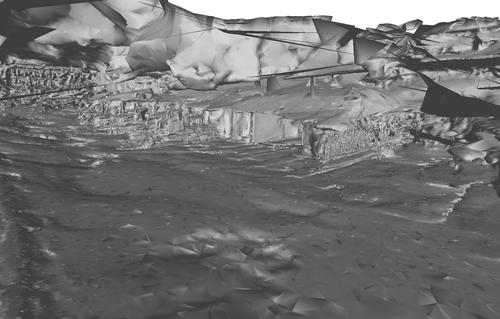} & 
            \includegraphics[clip=true, trim={0 0 0 0},width=0.19\textwidth]{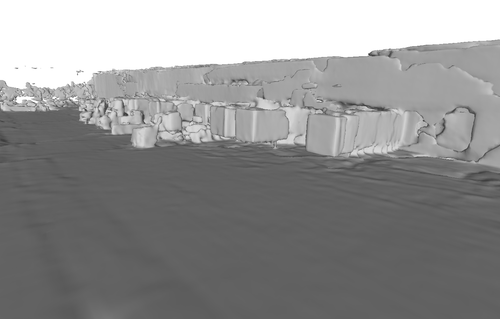} & 
            \includegraphics[clip=true, trim={0 0 0 0},width=0.19\textwidth]{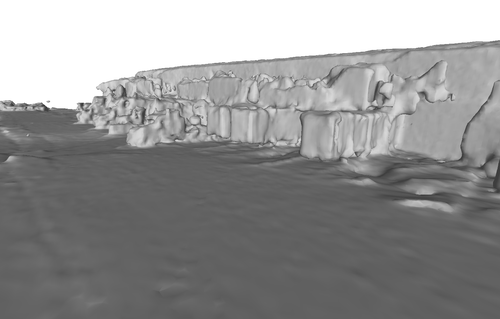} \\
        \end{tabular}
        \caption{nuScenes~\cite{nuscenes}} 
        \label{fig:qualitative/nuscenes}
    \end{subfigure}
    \hfill
    \begin{subfigure}{\textwidth}
        \centering
        \scriptsize
        \setlength{\tabcolsep}{0.002\linewidth}
        \renewcommand{\arraystretch}{0.8}
        \begin{tabular}{cccccc}
            \toprule
        
            &  COLMAP~\cite{schoenberger2016mvs} & OpenMVS~\cite{openmvs2020} &  GOF~\cite{Yu2024GOF} &StreetSurf~\cite{streetsurf} & ViiNeuS~(ours) \\ 

            \multirow{1}{*}[13.5mm]{\rotatebox[origin=c]{90}{Seq. 1319}}  &
            \includegraphics[clip=true, trim={0 7 0 0},width=0.19\textwidth]{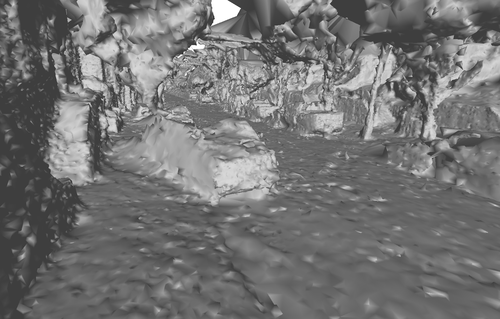} & 
            \includegraphics[clip=true, trim={0 0 0 0},width=0.19\textwidth]{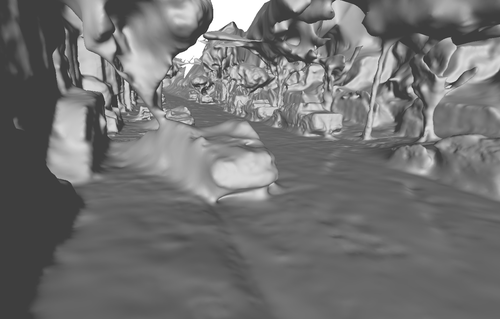} & 
            \includegraphics[clip=true, trim={0 0 0 0},width=0.19\textwidth]{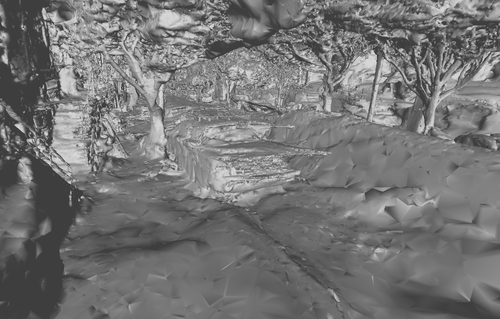} & 
            \includegraphics[clip=true, trim={0 0 0 0},width=0.19\textwidth]{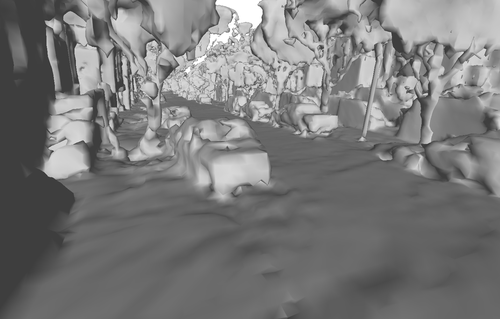} & 
            \includegraphics[clip=true, trim={0 0 0 0},width=0.19\textwidth]{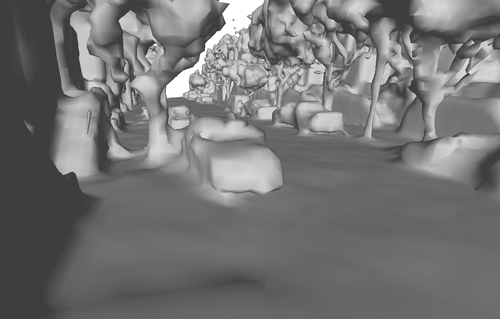} \\
            
            \multirow{1}{*}[13.5mm]{\rotatebox[origin=c]{90}{Seq. 1486}}  &
            \includegraphics[clip=true, trim={0 7 0 0},width=0.19\textwidth]{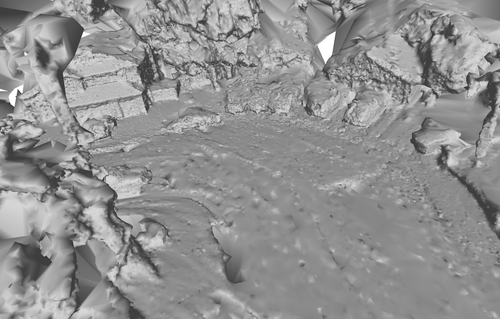} & 
            \includegraphics[clip=true, trim={0 0 0 0},width=0.19\textwidth]{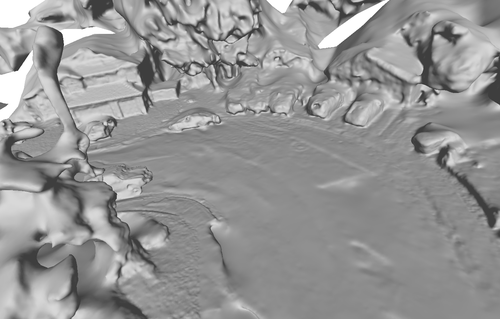} & 
            \includegraphics[clip=true, trim={0 0 0 0},width=0.19\textwidth]{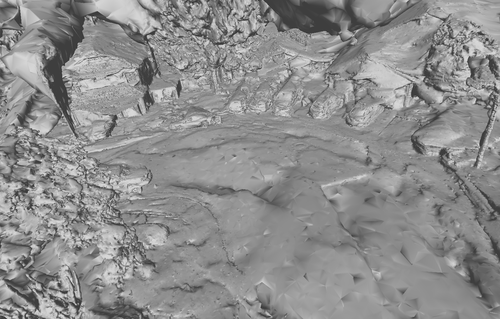} & 
            \includegraphics[clip=true, trim={0 0 0 0},width=0.19\textwidth]{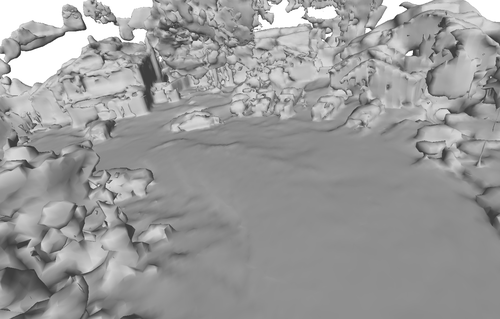} & 
            \includegraphics[clip=true, trim={0 0 0 0},width=0.19\textwidth]{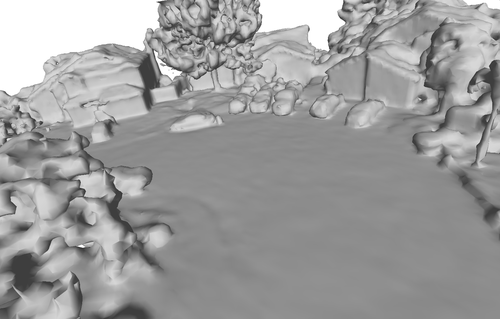}

        \end{tabular}
        \caption{Waymo Open Dataset~\cite{Waymo}} 
        \label{fig:qualitative/waymo}
    \end{subfigure}
  
      \caption{Qualitative experiments results on \protect\subref{fig:qualitative/kitti} KITTI-360s~\cite{Kitti}, \protect\subref{fig:qualitative/pandaset} Pandaset~\cite{pandaset}, \protect\subref{fig:qualitative/nuscenes} nuScenes~\cite{nuscenes} and \protect\subref{fig:qualitative/waymo} Waymo Open Dataset~\cite{Waymo}. %
        We compare our mesh extracted from our SDF to GOF, COLMAP, OpenMVS and StreetSurf meshes.}
   \label{fig:qualitative-pandaset}
\end{figure*}

We report in Fig.\ref{fig:qualitative-pandaset} additional qualitative results on nuScenes~\cite{nuscenes} and Waymo Open Dataset~\cite{Waymo}. We find that ViiNeuS reconstructs higher-quality surfaces compared to StreetSurf and can recover many scene details (see highlighted red-boxes on the figures).

\subsection{Ablation study}
\begin{figure*}[h] 
	\centering
	\scriptsize
	\setlength{\tabcolsep}{0.002\linewidth}
	\renewcommand{\arraystretch}{0.8}
	\begin{tabular}{cccc}

                \includegraphics[clip=true, trim={5 0 5 0},width=0.24\textwidth]{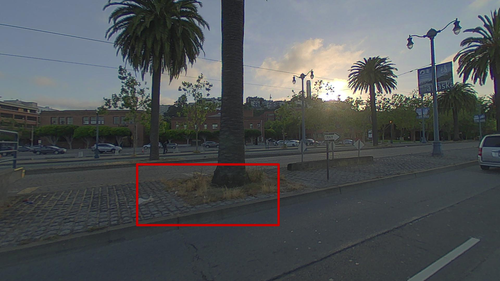} & 
            \includegraphics[clip=true, trim={5 0 5 0},width=0.24\textwidth]{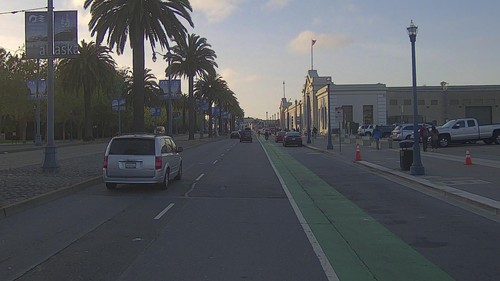} & 
        \includegraphics[clip=true, trim={0 0 0 0},width=0.24\textwidth]{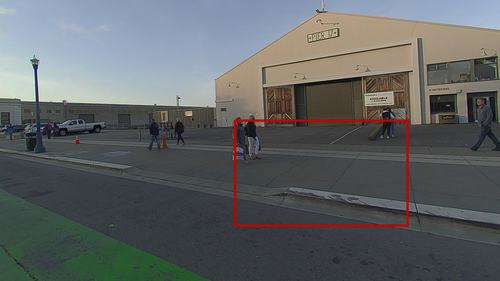} \\

         \textbf{(a)} Font Left RGB & \textbf{(b)} Front RGB  &  Front right RGB  \\
         \end{tabular}
        \begin{tabular}{ccccc}
        \multirow{1}{*}[15mm]{\rotatebox[origin=c]{90}{Random}}  &
        \includegraphics[clip=true, trim={0 0 0 0},width=0.24\textwidth]{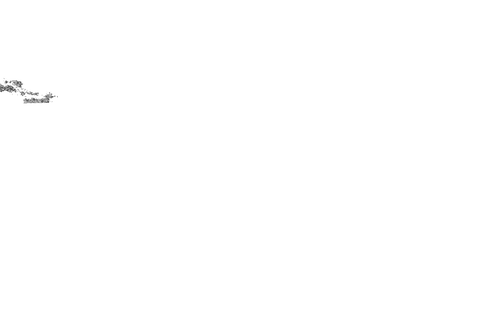} & 
            \includegraphics[clip=true, trim={0 0 0 0},width=0.24\textwidth]{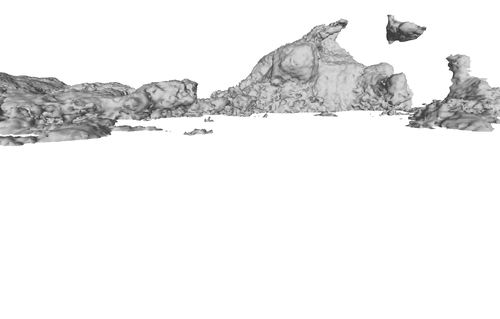} & 
        \includegraphics[clip=true, trim={0 0 0 0},width=0.24\textwidth]{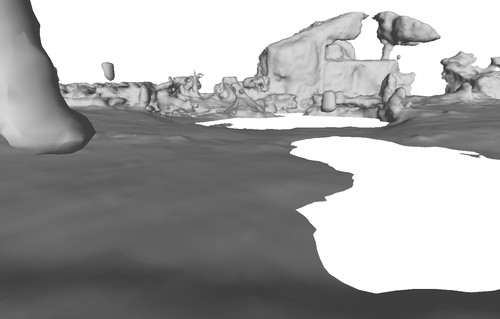} &
        \includegraphics[clip=true, trim={0 0 0 0},width=0.24\textwidth]{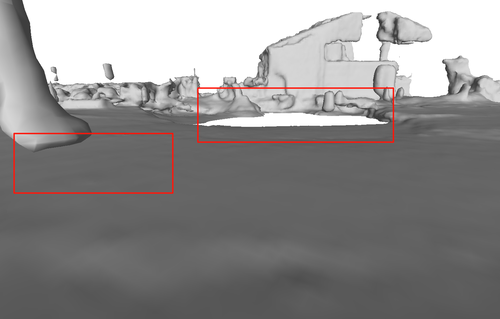} \\
        
        \multirow{1}{*}[15mm]{\rotatebox[origin=c]{90}{Prob-based}}  &
        \includegraphics[clip=true, trim={0 0 0 0},width=0.24\textwidth]{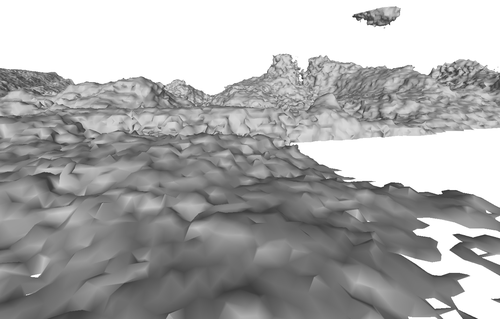} & 
            \includegraphics[clip=true, trim={0 0 0 0},width=0.24\textwidth]{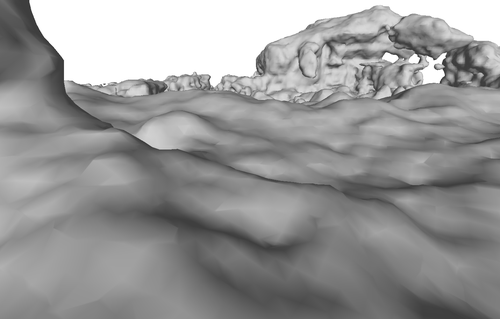} & 
        \includegraphics[clip=true, trim={0 0 0 0},width=0.24\textwidth]{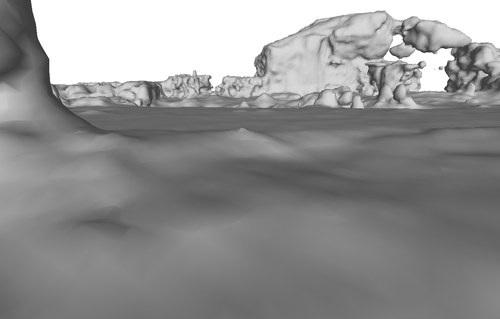} &
        \includegraphics[clip=true, trim={0 0 0 0},width=0.24\textwidth]{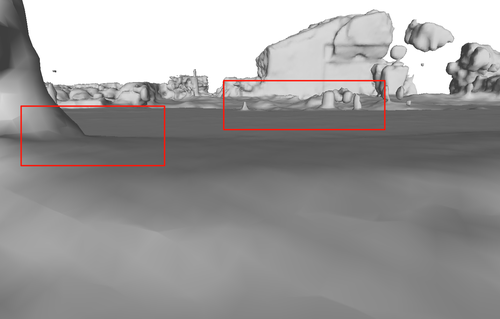} \\
        & \textbf{(a)} 1.2k steps & \textbf{(b)} 3.6k steps  &  \textbf{(c)} 6k steps & \textbf{(d)} 14k steps \\
		
	\end{tabular}
	\caption{Ablation study: we ablate the effect of random samples attribution compared to our probability-guided attribution introduced in section 3.3 of the paper. We show the rendered meshes results of the sequence 023 of Pandaset \cite{pandaset} at (a) 1.2k steps, (b) 3.6k steps, (c) 6k steps, and (d) 14k steps.
    }
	\label{fig:ablation-random}
\end{figure*}

We ablate the effect of random sample attribution and our proposed probability-based samples attribution. The qualitative results at various training steps are presented in ~Fig.\ref{fig:ablation-random}. The results demonstrate that ViiNeuS samples attribution strategy initially learns the coarse geometry of the scene during early training stages. While random sample attribution can approximate an accurate SDF representation by the end of the hybrid stage, it results in an incomplete mesh compared to the mesh generated using probability-based sample attribution.

\section{Applications}

\subsection{Textured mesh}
\begin{figure*}[h] 
	\centering
	\scriptsize
	\setlength{\tabcolsep}{0.002\linewidth}
	\renewcommand{\arraystretch}{0.8}
	\begin{tabular}{ccc}        
        \multirow{1}{*}[25mm]{\rotatebox[origin=c]{90}{ViiNeuS~(ours)}}  &
        \includegraphics[clip=true, trim={0 0 0 0},width=0.40\textwidth]{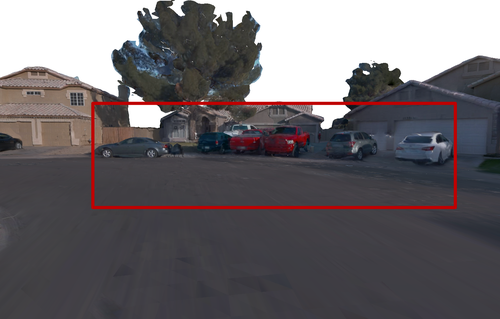} & 
            \includegraphics[clip=true, trim={0 0 0 0},width=0.40\textwidth]{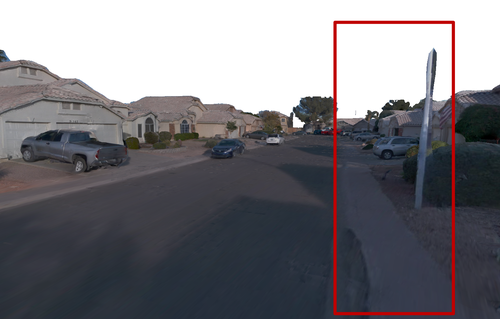}  \\
        \multirow{1}{*}[25mm]{\rotatebox[origin=c]{90}{StreetSurf~\cite{streetsurf}}}  &
        \includegraphics[clip=true, trim={0 0 0 0},width=0.40\textwidth]{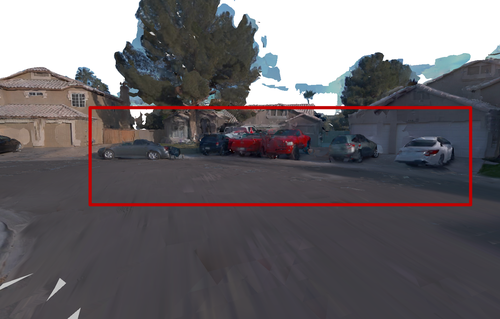} & 
        \includegraphics[clip=true, trim={0 0 0 0},width=0.40\textwidth]{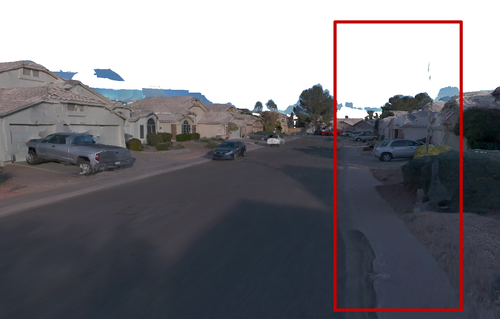}  \\
	\end{tabular}
	\caption{We use OpenMVS~\cite{openmvs2020} to assign texture to the outputted meshes. We compare our colored mesh to StreetSurf~\cite{streetsurf}, for the sequence 102751 from Waymo Open Dataset~\cite{Waymo}.
    }
	\label{fig:textured}
\end{figure*}

Due to the high-quality of ViiNeuS's reconstructed surfaces, we can leverage modern Multi-View Stereo (MVS) tools like OpenMVS~\cite{openmvs2020} to produce detailed and colorized representations of driving sequences. As shown in Fig.~\ref{fig:textured}, we find that ViiNeuS's textured mesh is more complete and accurate compared to StreetSurf's textured mesh.

\begin{figure}[tb] 
	\centering
	\scriptsize
	\setlength{\tabcolsep}{0.002\linewidth}
	\renewcommand{\arraystretch}{0.8}
	\begin{tabular}{cc}

        \includegraphics[clip=true, trim={5 0 5 0},width=0.49\columnwidth]{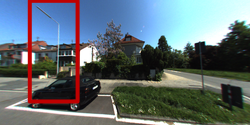} & 
            \includegraphics[clip=true, trim={5 0 5 0},width=0.49\columnwidth]{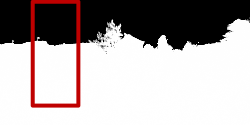} \\ 
            \textbf{(a)} RGB & \textbf{(b)} Predicted sky mask \\ 
        \includegraphics[clip=true, trim={0 0 0 0},width=0.49\columnwidth]{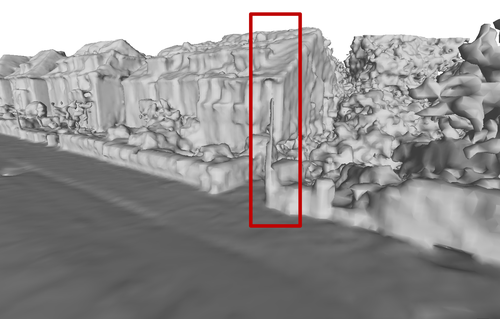} &
        \includegraphics[clip=true, trim={0 0 0 0},width=0.49\columnwidth]{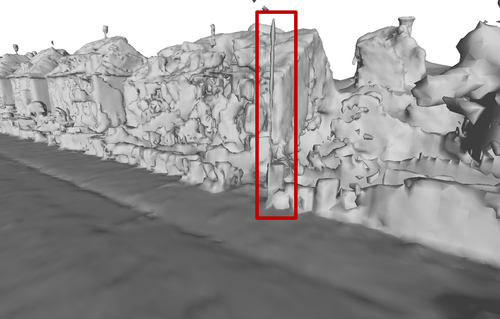} \\

         \textbf{(c)} ViiNeuS~(ours) & \textbf{(d)} StreetSurf~\cite{streetsurf} \\
		
	\end{tabular}
	\caption{Failure case of ViiNeuS for the sequence 31 from KITTI-360~\cite{Kitti}. We report the ground-truth RGB image and the predicted sky mask. In a different point-of-view than the GT RGB, we compare our generated SDF mesh to StreetSurf's mesh. 
    }
	\label{fig:failure-kitti}
\end{figure} 
\begin{figure}[tb] 
	\centering
	\scriptsize
	\setlength{\tabcolsep}{0.002\linewidth}
	\renewcommand{\arraystretch}{0.8}
	\begin{tabular}{cc}

        \includegraphics[clip=true, trim={5 0 5 0},width=0.49\columnwidth]{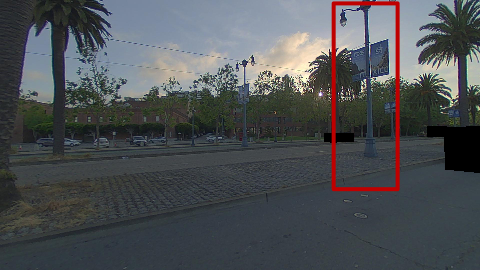} & 
            \includegraphics[clip=true, trim={5 0 5 0},width=0.49\columnwidth]{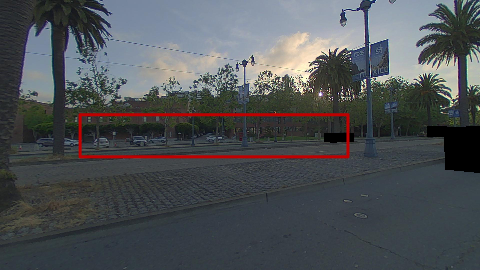} \\
            \textbf{(a)} RGB & \textbf{(b)} RGB highlighted \\
        \includegraphics[clip=true, trim={0 0 0 0},width=0.49\columnwidth]{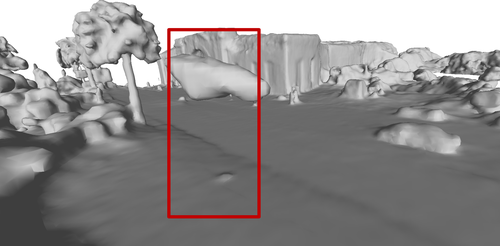} &
        \includegraphics[clip=true, trim={0 0 0 0},width=0.49\columnwidth]{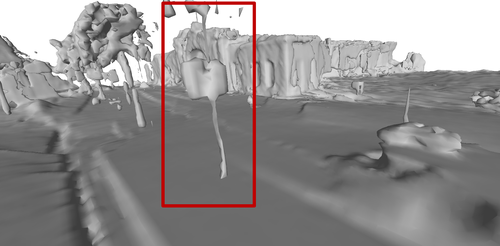} \\

           \textbf{(c)} ViiNeuS~(ours) & \textbf{(d)} StreetSurf~\cite{streetsurf} \\
		
	\end{tabular}
	\caption{Failure case of ViiNeuS for the sequence 23 from Pandaset~\cite{pandaset}. We report the ground-truth RGB image and the wide part of the sequence highlighted. In a different point-of-view than the GT RGB, we compare our generated SDF mesh to StreetSurf's mesh.
    }
	\label{fig:failure-pandaset}
\end{figure} 
\begin{figure}[tb] 
	\centering
	\scriptsize
	\setlength{\tabcolsep}{0.002\linewidth}
	\renewcommand{\arraystretch}{0.8}
	\begin{tabular}{cc}

        \includegraphics[clip=true, trim={0 0 0 0},width=0.49\columnwidth]{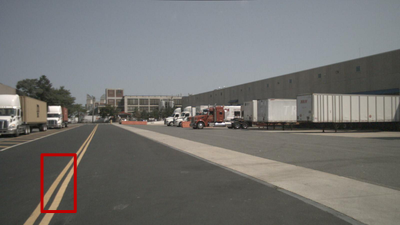} & 
            \includegraphics[clip=true, trim={5 0 5 0},width=0.49\columnwidth]{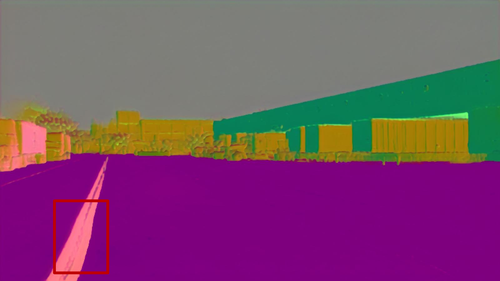} \\
            \textbf{(a)} RGB & \textbf{(b)} Monocular normal \\
        \includegraphics[clip=true, trim={0 0 0 0},width=0.49\columnwidth]{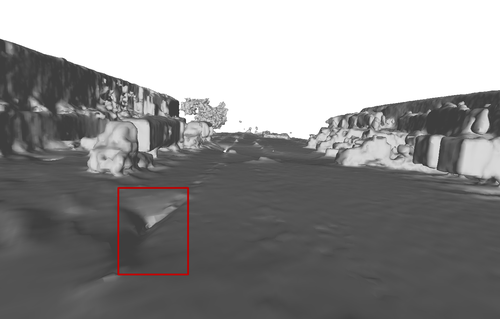}&
        \includegraphics[clip=true, trim={0 0 0 0},width=0.49\columnwidth]{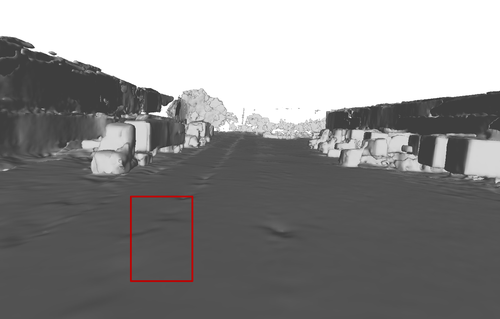} \\

            \textbf{(c)} ViiNeuS~(ours) & \textbf{(d)} StreetSurf~\cite{streetsurf} \\
		
	\end{tabular}
	\caption{Failure case of ViiNeuS for the sequence 664 from nuScenes~\cite{nuscenes}. We report the ground-truth RGB image and the monocular normal. In a different point-of-view than the GT RGB we compare our generated SDF mesh to StreetSurf's mesh.
    }
	\label{fig:failure-nuscenes}
\end{figure} 

\section{Limitations}
Although ViiNeuS's reconstructed surfaces are highly detailed and accurate, we find that our method can fail in three distinct scenarios: 

\begin{itemize}
    \item Disentangling fine details from the sky: unlike StreetSurf~\cite{streetsurf}, which models close range, far range, and sky separately, ViiNeuS separates only the sky from the other scene's modeling. However, ViiNeuS may struggle to distinguish fine details from the sky, particularly in cases where objects are thin, as shown in Fig.~\ref{fig:failure-kitti}.
    \item Fine details in wide sequences: ViiNeuS sometimes fails in accurately reconstructing scene details in wide and open sequences, such as scene 23 from Pandaset reported in Fig.~\ref{fig:failure-pandaset}.
    \item Inaccurate road reconstruction: for sequence 664 from nuScenes (see Fig.~\ref{fig:failure-nuscenes}), ViiNeuS faces challenges in reconstructing the road due to inaccuracies in the monocular normal prediction from Omnidata~\cite{eftekhar2021omnidata}. In comparison, StreetSurf~\cite{streetsurf} demonstrates more accurate road reconstruction, attributed to its road-surface initialization.  
\end{itemize}

\section{Supplementary video}
We show in the supplementary video ViiNeuS's meshes compared to GOF~\cite{Yu2024GOF} and StreetSurf~\cite{streetsurf} on one sequence from each of the four evaluated datasets. All meshes were visualized with Blender~\cite{blender} by animating the camera trajectory to generate the videos. GOF~\cite{Yu2024GOF} meshes are incomplete at the beginning of scenes and very noisy, as the method is designed for landmark reconstruction and does not address the challenges of driving sequences, such as low image overlap, off-centered regions of interest, the need to handle both close and far-range objects across a wide range of distances, and sky modeling. In addition, in the sky region and empty spaces that are commonly found in driving scenes, GOF tends to create triangles from noisy Gaussians. While the explicit 3DGS formulation is tailored for sparse scenes, it cannot effectively manage the inherent complexities introduced by the specific sensor configurations in driving sequences.

%% file: figures_tex/error_curves/error_curves.tex
\begin{figure*}[!t]
    \centering
    \setlength{\tabcolsep}{0.001\linewidth}
    \begin{tabular}{cccc}
        KITTI-360~\cite{Kitti} & Pandaset~\cite{pandaset} & Waymo~\cite{Waymo} & nuScenes~\cite{nuscenes} \\
         \includegraphics[height=.155\linewidth]{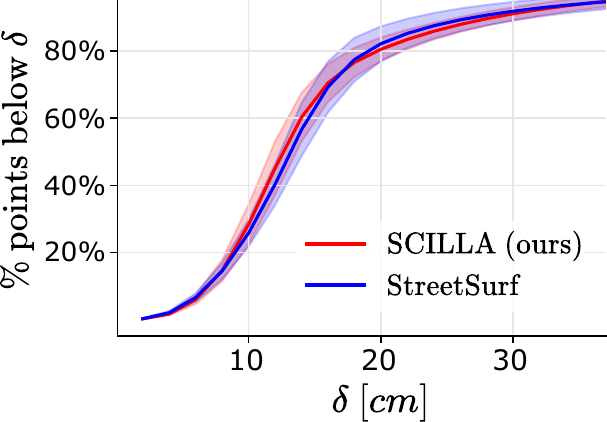}~~~~~~~~& 
         \includegraphics[height=.155\linewidth]{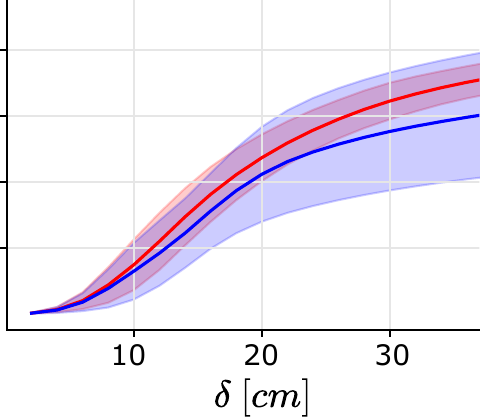}~~~~~~~~& 
         \includegraphics[height=.155\linewidth]{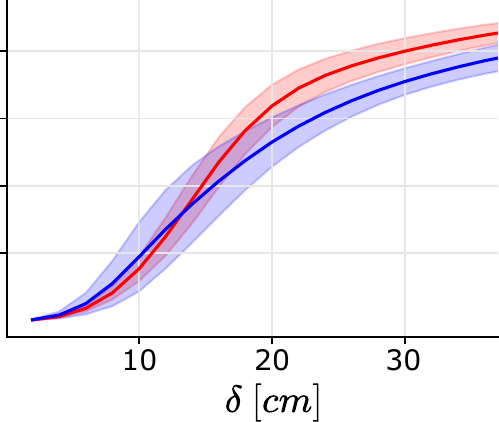}~~~~~~~~&
         \includegraphics[height=.155\linewidth]{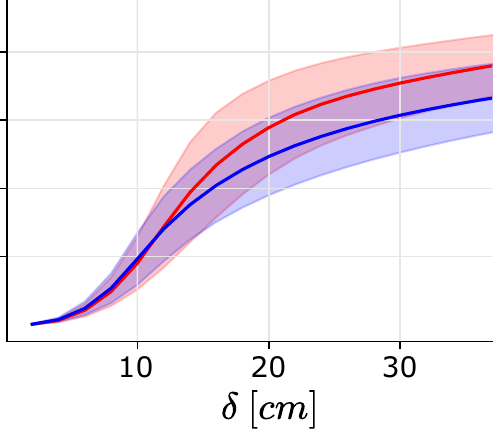}

    \end{tabular}
\vspace{-0.3cm}
\caption{Mean cumulative delta error for both \textcolor{red}{\method} and \textcolor{blue}{StreetSurf} computed across the four sequences from each dataset as detailed in Tab.~\textcolor{red}{1} of the main paper. Curves indicate the percentage of ground-truth points having an error distance to the closest predicted mesh triangle which is lower than a
given value. Light contours represent the standard deviation for each method. }
\label{fig:results}
\label{fig:distance-analysis}
\hfill
\vspace{-1mm}
\end{figure*}

%% file: main.bbl
\begin{thebibliography}{50}
\providecommand{\natexlab}[1]{#1}
\providecommand{\url}[1]{\texttt{#1}}
\expandafter\ifx\csname urlstyle\endcsname\relax
  \providecommand{\doi}[1]{doi: #1}\else
  \providecommand{\doi}{doi: \begingroup \urlstyle{rm}\Url}\fi

\bibitem[Barron et~al.(2022)Barron, Mildenhall, Verbin, Srinivasan, and Hedman]{mip-nerf-360}
Jonathan~T. Barron, Ben Mildenhall, Dor Verbin, Pratul~P. Srinivasan, and Peter Hedman.
\newblock Mip-nerf 360: Unbounded anti-aliased neural radiance fields.
\newblock In \emph{CVPR}, 2022.

\bibitem[Berger et~al.(2017)Berger, Tagliasacchi, Seversky, Alliez, Guennebaud, Levine, Sharf, and Silva]{Berger2017ASO}
Matthew Berger, Andrea Tagliasacchi, Lee~M. Seversky, Pierre Alliez, Ga{\"e}l Guennebaud, Joshua~A. Levine, Andrei Sharf, and Cl{\'a}udio~T. Silva.
\newblock A survey of surface reconstruction from point clouds.
\newblock \emph{Computer Graphics Forum}, 36, 2017.

\bibitem[Caesar et~al.(2020)Caesar, Bankiti, Lang, Vora, Liong, Xu, Krishnan, Pan, Baldan, and Beijbom]{nuscenes}
Holger Caesar, Varun Bankiti, Alex~H. Lang, Sourabh Vora, Venice~Erin Liong, Qiang Xu, Anush Krishnan, Yu Pan, Giancarlo Baldan, and Oscar Beijbom.
\newblock nuscenes: A multimodal dataset for autonomous driving.
\newblock In \emph{CVPR}, 2020.

\bibitem[Chen et~al.(2023)Chen, Li, and Lee]{chen2023neusg}
Hanlin Chen, Chen Li, and Gim~Hee Lee.
\newblock Neusg: Neural implicit surface reconstruction with 3d gaussian splatting guidance, 2023.

\bibitem[Cheng et~al.(2022)Cheng, Misra, Schwing, Kirillov, and Girdhar]{cheng2021mask2former}
Bowen Cheng, Ishan Misra, Alexander~G. Schwing, Alexander Kirillov, and Rohit Girdhar.
\newblock Masked-attention mask transformer for universal image segmentation.
\newblock In \emph{CVPR}, 2022.

\bibitem[Community(2018)]{blender}
Blender~Online Community.
\newblock \emph{Blender - a 3D modelling and rendering package}.
\newblock Blender Foundation, Stichting Blender Foundation, Amsterdam, 2018.

\bibitem[De~Bonet and Viola(1999)]{de1999poxels}
Jeremy~S De~Bonet and Paul Viola.
\newblock Poxels: Probabilistic voxelized volume reconstruction.
\newblock In \emph{ICCV}, 1999.

\bibitem[Eftekhar et~al.(2021)Eftekhar, Sax, Malik, and Zamir]{eftekhar2021omnidata}
Ainaz Eftekhar, Alexander Sax, Jitendra Malik, and Amir Zamir.
\newblock Omnidata: A scalable pipeline for making multi-task mid-level vision datasets from 3d scans.
\newblock In \emph{ICCV}, 2021.

\bibitem[Galliani et~al.(2016)Galliani, Lasinger, and Schindler]{galliani2016gipuma}
Silvano Galliani, Katrin Lasinger, and Konrad Schindler.
\newblock Gipuma: Massively parallel multi-view stereo reconstruction.
\newblock In \emph{ICCV}, 2016.

\bibitem[Gropp et~al.(2020)Gropp, Yariv, Haim, Atzmon, and Lipman]{icml2020_2086}
Amos Gropp, Lior Yariv, Niv Haim, Matan Atzmon, and Yaron Lipman.
\newblock Implicit geometric regularization for learning shapes.
\newblock In \emph{MLSYS}. 2020.

\bibitem[Gu\'edon and Lepetit(2024)]{Guedon_2024_CVPR}
Antoine Gu\'edon and Vincent Lepetit.
\newblock Sugar: Surface-aligned gaussian splatting for efficient 3d mesh reconstruction and high-quality mesh rendering.
\newblock In \emph{CVPR}, 2024.

\bibitem[Guo et~al.(2023{\natexlab{a}})Guo, Deng, Li, Bai, Shi, Wang, Ding, Wang, and Li]{streetsurf}
Jianfei Guo, Nianchen Deng, Xinyang Li, Yeqi Bai, Botian Shi, Chiyu Wang, Chenjing Ding, Dongliang Wang, and Yikang Li.
\newblock Streetsurf: Extending multi-view implicit surface reconstruction to street views, 2023{\natexlab{a}}.

\bibitem[Guo et~al.(2023{\natexlab{b}})Guo, Cao, Wang, He, Shan, Qie, and Zhang]{VMesh}
Yuan-Chen Guo, Yan-Pei Cao, Chen Wang, Yu He, Ying Shan, Xiaohu Qie, and Song-Hai Zhang.
\newblock Vmesh: Hybrid volume-mesh representation for efficient view synthesis.
\newblock In \emph{SIGGRAPH Asia}. ACM, 2023{\natexlab{b}}.

\bibitem[Herau et~al.(2023)Herau, Piasco, Bennehar, Roldão, Tsishkou, Migniot, Vasseur, and Demonceaux]{Herau_2023}
Quentin Herau, Nathan Piasco, Moussab Bennehar, Luis Roldão, Dzmitry Tsishkou, Cyrille Migniot, Pascal Vasseur, and Cédric Demonceaux.
\newblock Moisst: Multimodal optimization of implicit scene for spatiotemporal calibration.
\newblock In \emph{IROS}, 2023.

\bibitem[Huang et~al.(2024{\natexlab{a}})Huang, Yu, Chen, Geiger, and Gao]{Huang2DGS2024}
Binbin Huang, Zehao Yu, Anpei Chen, Andreas Geiger, and Shenghua Gao.
\newblock 2d gaussian splatting for geometrically accurate radiance fields.
\newblock In \emph{SIGGRAPH 2024 Conference Papers}. Association for Computing Machinery, 2024{\natexlab{a}}.

\bibitem[Huang et~al.(2024{\natexlab{b}})Huang, Liang, Zhang, Lin, and Jia]{huang2024sur2f}
Zhangjin Huang, Zhihao Liang, Haojie Zhang, Yangkai Lin, and Kui Jia.
\newblock Sur2f: A hybrid representation for high-quality and efficient surface reconstruction from multi-view images, 2024{\natexlab{b}}.

\bibitem[Jancosek and Pajdla(2014)]{Jancosek2014ExploitingVI}
Michal Jancosek and Tom{\'a}s Pajdla.
\newblock Exploiting visibility information in surface reconstruction to preserve weakly supported surfaces.
\newblock \emph{International Scholarly Research Notices}, 2014, 2014.

\bibitem[Kazhdan et~al.(2006)Kazhdan, Bolitho, and Hoppe]{kazhdan2006poisson}
Michael Kazhdan, Matthew Bolitho, and Hugues Hoppe.
\newblock Poisson surface reconstruction.
\newblock In \emph{Eurographics Symp. Geometry Processing}, page 61–70, 2006.

\bibitem[Kerbl et~al.(2023)Kerbl, Kopanas, Leimk{\"u}hler, and Drettakis]{kerbl3Dgaussians}
Bernhard Kerbl, Georgios Kopanas, Thomas Leimk{\"u}hler, and George Drettakis.
\newblock 3d gaussian splatting for real-time radiance field rendering.
\newblock \emph{ACM Transactions on Graphics}, 42\penalty0 (4), 2023.

\bibitem[Li et~al.(2023)Li, M\"uller, Evans, Taylor, Unberath, Liu, and Lin]{li2023neuralangelo}
Zhaoshuo Li, Thomas M\"uller, Alex Evans, Russell~H Taylor, Mathias Unberath, Ming-Yu Liu, and Chen-Hsuan Lin.
\newblock Neuralangelo: High-fidelity neural surface reconstruction.
\newblock In \emph{CVPR}, 2023.

\bibitem[Liao et~al.(2022)Liao, Xie, and Geiger]{Kitti}
Yiyi Liao, Jun Xie, and Andreas Geiger.
\newblock {KITTI}-360: A novel dataset and benchmarks for urban scene understanding in 2d and 3d.
\newblock \emph{PAMI}, 2022.

\bibitem[Lorensen and Cline(1987)]{10.1145/37401.37422}
William~E. Lorensen and Harvey~E. Cline.
\newblock Marching cubes: A high resolution 3d surface construction algorithm.
\newblock \emph{ACM Transactions on Graphics}, 21\penalty0 (4):\penalty0 163--169, 1987.

\bibitem[Malleson et~al.()Malleson, Guillemaut, and Hilton]{Malleson20193DRF}
Charles Malleson, Jean-Yves Guillemaut, and Adrian Hilton.
\newblock 3d reconstruction from rgb-d data.
\newblock \emph{RGB-D Image Analysis and Processing}.

\bibitem[Mildenhall et~al.(2020)Mildenhall, Srinivasan, Tancik, Barron, Ramamoorthi, and Ng]{2020nerf}
Ben Mildenhall, Pratul~P. Srinivasan, Matthew Tancik, Jonathan~T. Barron, Ravi Ramamoorthi, and Ren Ng.
\newblock Nerf: Representing scenes as neural radiance fields for view synthesis.
\newblock In \emph{ECCV}, 2020.

\bibitem[Moulon et~al.(2016{\natexlab{a}})Moulon, Monasse, Perrot, and Marlet]{openMVG}
Pierre Moulon, Pascal Monasse, Romuald Perrot, and Renaud Marlet.
\newblock Open{MVG}: Open multiple view geometry.
\newblock In \emph{International Workshop on Reproducible Research in Pattern Recognition}, pages 60--74. Springer, 2016{\natexlab{a}}.

\bibitem[Moulon et~al.(2016{\natexlab{b}})Moulon, Monasse, Perrot, and Marlet]{openmvs2020}
Pierre Moulon, Pascal Monasse, Romuald Perrot, and Renaud Marlet.
\newblock Open{MVG}: Open multiple view geometry.
\newblock In \emph{International Workshop on Reproducible Research in Pattern Recognition}, 2016{\natexlab{b}}.

\bibitem[M\"uller et~al.(2022)M\"uller, Evans, Schied, and Keller]{mueller2022instant}
Thomas M\"uller, Alex Evans, Christoph Schied, and Alexander Keller.
\newblock Instant neural graphics primitives with a multiresolution hash encoding.
\newblock \emph{ACM Trans. Graph.}, 41\penalty0 (4):\penalty0 102:1--102:15, 2022.

\bibitem[Paschalidou et~al.(2019)Paschalidou, Ulusoy, Schmitt, van Gool, and Geiger]{paschalidou2019raynet}
Despoina Paschalidou, Ali~Osman Ulusoy, Carolin Schmitt, Luc van Gool, and Andreas Geiger.
\newblock Raynet: Learning volumetric 3d reconstruction with ray potentials.
\newblock In \emph{CVPR}, 2019.

\bibitem[Rematas et~al.(2022)Rematas, Liu, Srinivasan, Barron, Tagliasacchi, Funkhouser, and Ferrari]{urban-radiance-fields}
Konstantinos Rematas, Andrew Liu, Pratul~P. Srinivasan, Jonathan~T. Barron, Andrea Tagliasacchi, Thomas Funkhouser, and Vittorio Ferrari.
\newblock Urban radiance fields.
\newblock In \emph{CVPR}, 2022.

\bibitem[Rold{\~a}o et~al.(2019)Rold{\~a}o, de~Charette, and Verroust-Blondet]{Roldo20193DSR}
Luis Rold{\~a}o, Raoul de Charette, and Anne Verroust-Blondet.
\newblock 3d surface reconstruction from voxel-based lidar data.
\newblock \emph{(ITSC)}, pages 2681--2686, 2019.

\bibitem[Sch{\"o}nberger et~al.(2016)Sch{\"o}nberger, Zheng, Frahm, and Pollefeys]{schonberger2016pixelwise}
Johannes~L Sch{\"o}nberger, Enliang Zheng, Jan-Michael Frahm, and Marc Pollefeys.
\newblock Pixelwise view selection for unstructured multi-view stereo.
\newblock In \emph{ECCV}, 2016.

\bibitem[Sch\"{o}nberger et~al.(2016)Sch\"{o}nberger, Zheng, Pollefeys, and Frahm]{schoenberger2016mvs}
Johannes~Lutz Sch\"{o}nberger, Enliang Zheng, Marc Pollefeys, and Jan-Michael Frahm.
\newblock Pixelwise view selection for unstructured multi-view stereo.
\newblock In \emph{ECCV}, 2016.

\bibitem[Stereopsis(2010)]{stereopsis2010accurate}
Robust~Multiview Stereopsis.
\newblock Accurate, dense, and robust multiview stereopsis.
\newblock \emph{TPAMI}, 32\penalty0 (8):\penalty0 1362--1376, 2010.

\bibitem[Sun et~al.(2020)Sun, Kretzschmar, Dotiwalla, Chouard, Patnaik, Tsui, Guo, Zhou, Chai, Caine, Vasudevan, Han, Ngiam, Zhao, Timofeev, Ettinger, Krivokon, Gao, Joshi, Zhang, Shlens, Chen, and Anguelov]{Waymo}
Pei Sun, Henrik Kretzschmar, Xerxes Dotiwalla, Aurelien Chouard, Vijaysai Patnaik, Paul Tsui, James Guo, Yin Zhou, Yuning Chai, Benjamin Caine, Vijay Vasudevan, Wei Han, Jiquan Ngiam, Hang Zhao, Aleksei Timofeev, Scott Ettinger, Maxim Krivokon, Amy Gao, Aditya Joshi, Yu Zhang, Jonathon Shlens, Zhifeng Chen, and Dragomir Anguelov.
\newblock Scalability in perception for autonomous driving: Waymo open dataset.
\newblock In \emph{CVPR}, 2020.

\bibitem[Tang et~al.(2023)Tang, Zhou, Chen, Hu, Ding, Wang, and Zeng]{tang2023delicate}
Jiaxiang Tang, Hang Zhou, Xiaokang Chen, Tianshu Hu, Errui Ding, Jingdong Wang, and Gang Zeng.
\newblock Delicate textured mesh recovery from nerf via adaptive surface refinement.
\newblock In \emph{ICCV}, 2023.

\bibitem[Tulsiani et~al.(2017)Tulsiani, Zhou, Efros, and Malik]{tulsiani2017multiview}
Shubham Tulsiani, Tinghui Zhou, Alexei~A. Efros, and Jitendra Malik.
\newblock Multi-view supervision for single-view reconstruction via differentiable ray consistency.
\newblock In \emph{CVPR}, 2017.

\bibitem[Ulusoy et~al.(2015)Ulusoy, Geiger, and Black]{7335464}
Ali~Osman Ulusoy, Andreas Geiger, and Michael~J. Black.
\newblock Towards probabilistic volumetric reconstruction using ray potentials.
\newblock In \emph{3DV}, 2015.

\bibitem[Wang et~al.(2024)Wang, Louys, Piasco, Bennehar, Roldãao, and Tsishkou]{wang2023planerf}
Fusang Wang, Arnaud Louys, Nathan Piasco, Moussab Bennehar, Luis Roldãao, and Dzmitry Tsishkou.
\newblock Planerf: Svd unsupervised 3d plane regularization for nerf large-scale urban scene reconstruction.
\newblock In \emph{3DV}, 2024.

\bibitem[Wang et~al.(2021)Wang, Liu, Liu, Theobalt, Komura, and Wang]{NeuS}
Peng Wang, Lingjie Liu, Yuan Liu, Christian Theobalt, Taku Komura, and Wenping Wang.
\newblock Neus: Learning neural implicit surfaces by volume rendering for multi-view reconstruction.
\newblock In \emph{NeurIPS}, 2021.

\bibitem[Wang et~al.(2023{\natexlab{a}})Wang, Han, Habermann, Daniilidis, Theobalt, and Liu]{neus2}
Yiming Wang, Qin Han, Marc Habermann, Kostas Daniilidis, Christian Theobalt, and Lingjie Liu.
\newblock Neus2: Fast learning of neural implicit surfaces for multi-view reconstruction.
\newblock In \emph{ICCV}, 2023{\natexlab{a}}.

\bibitem[Wang et~al.(2023{\natexlab{b}})Wang, Shen, Gao, Huang, Munkberg, Hasselgren, Gojcic, Chen, and Fidler]{fegr}
Zian Wang, Tianchang Shen, Jun Gao, Shengyu Huang, Jacob Munkberg, Jon Hasselgren, Zan Gojcic, Wenzheng Chen, and Sanja Fidler.
\newblock Neural fields meet explicit geometric representation for inverse rendering of urban scenes.
\newblock In \emph{CVPR}, 2023{\natexlab{b}}.

\bibitem[Wang et~al.(2023{\natexlab{c}})Wang, Shen, Nimier-David, Sharp, Gao, Keller, Fidler, Müller, and Gojcic]{wang2023adaptive}
Zian Wang, Tianchang Shen, Merlin Nimier-David, Nicholas Sharp, Jun Gao, Alexander Keller, Sanja Fidler, Thomas Müller, and Zan Gojcic.
\newblock Adaptive shells for efficient neural radiance field rendering.
\newblock In \emph{SIGGRAPH Asia}, 2023{\natexlab{c}}.

\bibitem[Xiao et~al.(2021)Xiao, Shao, Hao, Zhang, Chai, Jiao, Li, Wu, Sun, Jiang, et~al.]{pandaset}
Pengchuan Xiao, Zhenlei Shao, Steven Hao, Zishuo Zhang, Xiaolin Chai, Judy Jiao, Zesong Li, Jian Wu, Kai Sun, Kun Jiang, et~al.
\newblock Pandaset: Advanced sensor suite dataset for autonomous driving.
\newblock In \emph{ITSC}, 2021.

\bibitem[Yariv et~al.(2020)Yariv, Kasten, Moran, Galun, Atzmon, Ronen, and Lipman]{yariv2020multiview}
Lior Yariv, Yoni Kasten, Dror Moran, Meirav Galun, Matan Atzmon, Basri Ronen, and Yaron Lipman.
\newblock Multiview neural surface reconstruction by disentangling geometry and appearance.
\newblock \emph{Advances in Neural Information Processing Systems}, 33:\penalty0 2492--2502, 2020.

\bibitem[Yariv et~al.(2021)Yariv, Gu, Kasten, and Lipman]{VolSDF}
Lior Yariv, Jiatao Gu, Yoni Kasten, and Yaron Lipman.
\newblock Volume rendering of neural implicit surfaces.
\newblock In \emph{NeurIPS}, 2021.

\bibitem[Yu et~al.(2022)Yu, Peng, Niemeyer, Sattler, and Geiger]{Yu2022MonoSDF}
Zehao Yu, Songyou Peng, Michael Niemeyer, Torsten Sattler, and Andreas Geiger.
\newblock Monosdf: Exploring monocular geometric cues for neural implicit surface reconstruction.
\newblock \emph{NeurIPS}, 2022.

\bibitem[Yu et~al.(2024)Yu, Sattler, and Geiger]{Yu2024GOF}
Zehao Yu, Torsten Sattler, and Andreas Geiger.
\newblock Gaussian opacity fields: Efficient adaptive surface reconstruction in unbounded scenes.
\newblock \emph{ACM Transactions on Graphics}, 2024.

\bibitem[Zhang et~al.(2023)Zhang, Yao, Li, Liu, Fang, McKinnon, Tsin, and Quan]{zhang2023neilf}
Jingyang Zhang, Yao Yao, Shiwei Li, Jingbo Liu, Tian Fang, David McKinnon, Yanghai Tsin, and Long Quan.
\newblock Neilf++: Inter-reflectable light fields for geometry and material estimation.
\newblock In \emph{ICCV}, 2023.

\bibitem[Zhu et~al.(2023)Zhu, Huo, Ye, Luan, Li, Xi, Wang, Tang, Hua, Bao, and Wang]{Zhu_2023_CVPR}
Jingsen Zhu, Yuchi Huo, Qi Ye, Fujun Luan, Jifan Li, Dianbing Xi, Lisha Wang, Rui Tang, Wei Hua, Hujun Bao, and Rui Wang.
\newblock I2-sdf: Intrinsic indoor scene reconstruction and editing via raytracing in neural sdfs.
\newblock In \emph{CVPR}, 2023.

\bibitem[Zollh{\"o}fer et~al.(2018)Zollh{\"o}fer, Stotko, G{\"o}rlitz, Theobalt, Nie{\ss}ner, Klein, and Kolb]{Zollhfer2018StateOT}
Michael Zollh{\"o}fer, Patrick Stotko, Andreas G{\"o}rlitz, Christian Theobalt, Matthias Nie{\ss}ner, R. Klein, and Andreas Kolb.
\newblock State of the art on 3d reconstruction with rgb‐d cameras.
\newblock \emph{Computer Graphics Forum}, 37, 2018.

\end{thebibliography}
